%% file: main.tex
\pdfoutput=1

\documentclass[11pt]{article}

\usepackage{ACL2023}

\usepackage[T1]{fontenc}
\usepackage[utf8]{inputenc}
\usepackage{times}
\usepackage{latexsym}
\usepackage{inconsolata}
\usepackage{microtype}

\usepackage{amssymb}   %
\usepackage{amsmath}   %
\usepackage{amsfonts}  %
\usepackage{amsthm}    %
\usepackage{latexsym}  %
\usepackage{stmaryrd}  %

\usepackage{soul}      %

\usepackage{graphicx}  %
\usepackage{subcaption}
\usepackage{color}     %
\usepackage{xcolor}    %
\usepackage{adjustbox} %
\usepackage{url}

\usepackage{url}       %
\usepackage{hyperref}  %

\usepackage{array}     %
\usepackage{multirow}  %
\usepackage{multicol}  %
\usepackage{hhline}    %
\usepackage{dcolumn}   %
\usepackage{booktabs}  %

\usepackage{tikz}      %
\usepackage{pgfplots}  %
\pgfplotsset{compat=1.16}

\title{RARR: Researching and Revising What Language Models Say, \\ Using Language Models}

\newcommand*\samethanks[1][\value{footnote}]{\footnotemark[#1]}

\author{
\textbf{Luyu Gao}$^{1\diamond}$\thanks{\enskip Lead contributors. Please see Contributions section for details. $^\diamond$Work done during an internship at Google Research.} \quad
\textbf{Zhuyun Dai}$^2$\samethanks[1] \quad
\textbf{Panupong Pasupat}$^2$\samethanks[1] \quad
\textbf{Anthony Chen}$^{3\diamond}$\samethanks[1] \quad
\\
\textbf{Arun Tejasvi Chaganty}$^2$\samethanks[1] \quad
\textbf{Yicheng Fan}$^2$\samethanks[1] \quad
\textbf{Vincent Y. Zhao}$^2$ \quad
\textbf{Ni Lao}$^2$ \quad
\\
\textbf{Hongrae Lee}$^2$ \quad
\textbf{Da-Cheng Juan}$^2$ \quad
\textbf{Kelvin Guu}$^2$\samethanks[1]
\\[.5em]
$^1$Carnegie Mellon University, $^2$Google Research, $^3$UC Irvine
\\
\small{\texttt{luyug@cs.cmu.edu\qquad anthony.chen@uci.edu}}
\\
\small{\texttt{\{zhuyundai,ppasupat,arunchaganty,yichengfan,vzhao,nlao,hrlee,dacheng,kguu\}@google.com}}
}

\date{}

\begin{document}

\maketitle

\input{macros}

\begin{abstract}
\input{abstract}
\end{abstract}

\input{0_intro}
\input{1_task}
\input{2_approach}
\input{3_related}

\input{4_results}

\input{5_discussion}

\input{6_contributions}

\bibliography{custom}
\bibliographystyle{acl_natbib}

\appendix
\input{7_appendix}

\end{document}

%% file: macros.tex
\newcommand{\approach}{RARR}
\newcommand{\tigeremoji}[0]{\includegraphics[height=.02\textwidth]{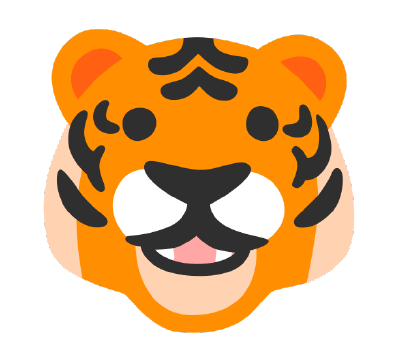}}
\newcommand{\approachfull}{\textit{Retrofit Attribution using Research and Revision}}

\newcommand{\kg}[1]{{\color{red}{\bf{[kguu]}} \emph{#1}}}
\newcommand{\nl}[1]{{\color{olive}{\bf{[nlao]}} \emph{#1}}}
\newcommand{\luyu}[1]{{\color{Emerald}{\bf{[luyu]}} \emph{#1}}}
\newcommand{\ac}[1]{{\color{blue}{\bf{[arun]}} \emph{#1}}}
\newcommand{\ice}[1]{{\color{cyan}{\bf{[ice]}} \emph{#1}}}

\newcommand{\zhuyun}[1]{{\color{ForestGreen}{\bf{[zhuyun]}} \emph{#1}}}

\newcommand{\edit}[2]{\st{#1} {\color{blue}{#2}}}

\newcommand{\taskName}{\textit{Editing for Attribution}}
\newcommand{\modelName}{\approachfull}

\newcommand{\familyName}{Text Generation Model}%
\newcommand{\fName}{TGM} %

\newcommand{\e}{e} %

\newcommand{\attrais}{\text{Attr}_\text{AIS}}
\newcommand{\attrauto}{\text{Attr}_\text{auto}}
\newcommand{\presintent}{\text{Pres}_\text{intent}}
\newcommand{\preslev}{\text{Pres}_\text{Lev}}
\newcommand{\prescomb}{\text{Pres}_\text{comb}}

\newcommand{\MN}[1]{\multicolumn{1}{c}{\footnotesize #1}}
\newcommand{\cmpR}[1]{\multicolumn{2}{@{}p{1.0\columnwidth}@{}}{#1}}

\newcommand{\HM}{\text{F1}_\text{AP}}

\hyphenation{SummEval}
\hyphenation{StrategyQA}

\definecolor{BrickRed}{RGB}{170, 74, 68}
\definecolor{DarkOrchid}{RGB}{153, 50, 204}

%% file: abstract.tex
Language models (LMs) now excel at many tasks such as %
question answering, reasoning, and dialog.
However, they sometimes generate unsupported or misleading content.
A user cannot easily determine whether their outputs are trustworthy or not, because most LMs do not have any built-in mechanism for \emph{attribution} to external evidence.
To enable attribution while still preserving all the powerful advantages of recent generation models, we propose \approach{} (\approachfull{}), a system that 1) automatically finds attribution for the output of any text generation model, and 2) post-edits the output to fix unsupported content while preserving the original output as much as possible.
When applied to the output of several state-of-the-art LMs on a diverse set of generation tasks, we find that \approach{} significantly improves attribution while otherwise preserving the original input to a much greater degree than previously explored edit models. Furthermore, the implementation of \approach{} requires only a handful of training examples, a large language model, and standard web search.\footnote{We release open-source implementations of \approach{}, the evaluation pipeline, and the evaluation sets at \url{https://github.com/anthonywchen/RARR}.}

%% file: 0_intro.tex
\input{fig_fate_api}
\section{Introduction} \label{sec:intro}
Generative language models (LMs) and other text generation models are now the backbone of many AI systems. For example, large language models can perform multi-step reasoning \cite{nye2021show, Wei2022COT}, generate plans \cite{Ahn2022SayCan}, use tools and APIs \cite{shin2021constrained,Thoppilan2022LaMDA}, and answer open-domain questions \cite{petroni2019language,roberts2020much}.

Despite these incredible advances, state-of-the-art LMs still frequently produce biased, misleading, or unsupported content, colloquially called ``hallucinations'' \cite{maynez2020faithfulness,menick2022teaching}. To make LMs more trustworthy, we want to justify each generation by an \emph{attribution report} \cite{Rashkin2021AIS,Bohnet2022AttributedQA} that contains supporting evidence from trusted sources (e.g., encyclopedia or articles) where appropriate.

Most existing LMs, such as those based on sequence-to-sequence architectures, lack a built-in mechanism for attribution. Even \emph{retrieval-augmented models} \cite{gu2020realm,lewis2020rag}, which retrieve relevant documents and then condition on them to generate text, still do not guarantee attribution.
Prior work has shown that retrieval-augmented models generate text that either includes additional information outside the retrieved documents \cite{dziri2022origin}, ignores the documents altogether \cite{krishna2021hurdles}, or even contradicts the documents \cite{longpre2021entity}. In fact, occasionally ignoring the retrievals can make the models more robust to bad retrievals \cite{khandelwal2019generalization}, illustrating that end-task performance and attribution are not always aligned.

Instead of constraining LMs to generate attributed text, we propose a model-agnostic approach to improve the attribution of any existing LM: \emph{\approachfull{}} (\approach{}).
The approach is inspired by works on fact-checking\footnote{In this paper, we generally avoid the term ``fact-checking'' other than to reference relevant literature, because we only address attribution, and attribution does not entail correctness. Even if a claim is attributed to a particular source, it does not guarantee that the source is ``correct'' \cite{menick2022teaching}.} where simple research-and-revise workflows are effective at attributing or correcting unattributed claims made by humans \cite{thorne2018fever,schuster2021get,thorne2021evidence}.
As shown in Figure~\ref{fig:api}, after generating text with the LM, \approach{} does \emph{research} to retrieve relevant evidence, and then \emph{revises} the text to make it consistent with the evidence while preserving qualities like style or structure, enabling the revised text to be seamlessly used in place of the original. \approach{} can be viewed as a retrieval-augmented model where retrieval happens \emph{after} generation rather than before. This allows \approach{} to stand on the shoulders of giant LMs without having to modify them to support attribution.

In our effort to expand the scope of Research \& Revision models to handle the output of arbitrary LMs, we make the following contributions.
First, we formalize the \taskName{} \textbf{task} and propose new \textbf{metrics} that evaluate revision models not just on their ability to produce well-attributed revisions, but also on their ability to otherwise \emph{preserve} original properties of the text.
Second, we use these metrics to \textbf{benchmark} how existing revision models perform on various types of LM outputs such as knowledge-intensive statements, reasoning chains, and dialog responses.
Finally, we find that existing revision models do not always generalize across many tasks (and were not originally intended to), and therefore propose a new research-and-revise \textbf{model} that leverages the power of few-shot prompting in large language models to robustly generalize across domains.

%% file: fig_fate_api.tex
\begin{figure}[!t]
\centering
\includegraphics[width=.9\columnwidth]{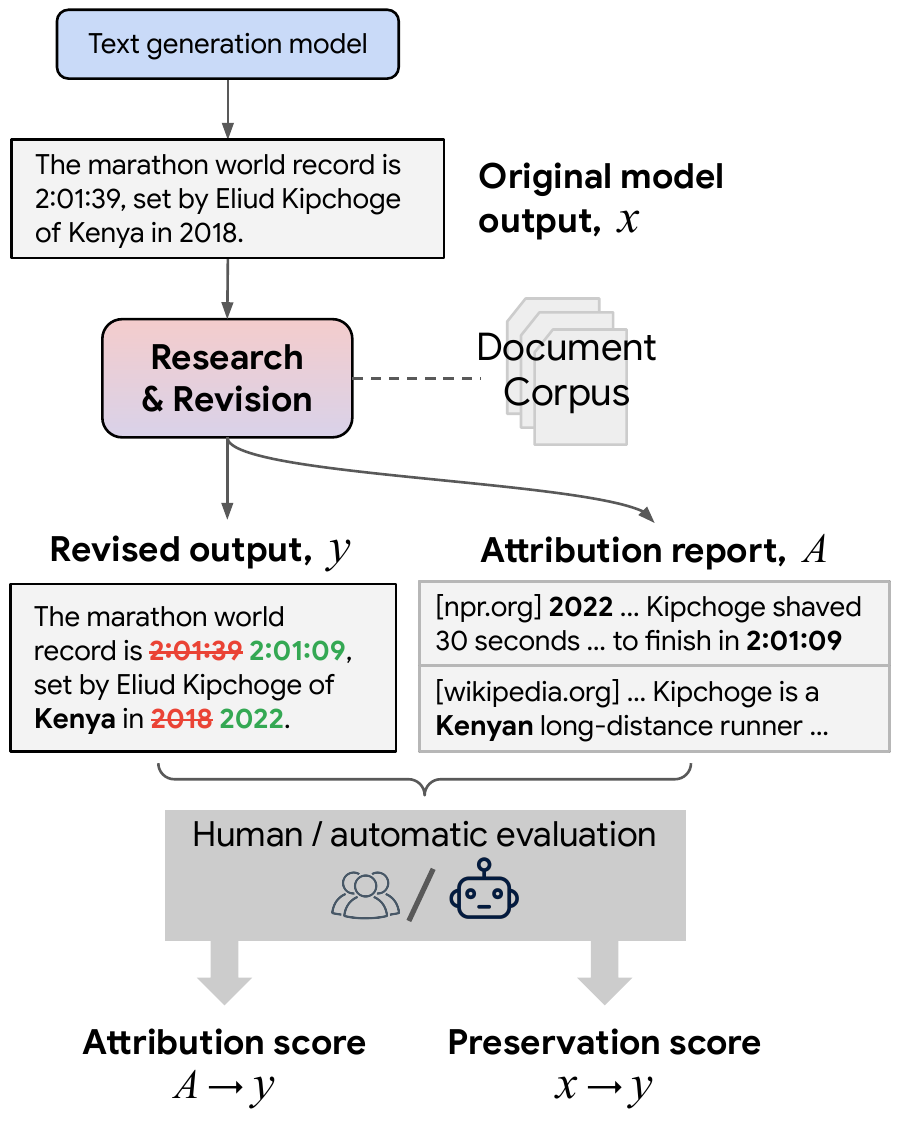}
\caption{
\label{fig:api}
\textbf{The \taskName{} task.}
The input $x$ is a text passage produced by a generation model.
Our \emph{Research \& Revision} model %
outputs an attribution report $A$ containing retrieved evidence snippets, along with a revision $y$ whose content can be \emph{attributed} to the evidence in $A$ while \emph{preserving} other properties of $x$ such as style or structure.
}
\end{figure}

%% file: 1_task.tex
\input{fig_overview}
\section{Task formulation}\label{sec:task}

We propose the task of \textit{\taskName} as follows. As Figure~\ref{fig:api} shows, the input to the system is a text passage $x$ produced by a generation model. %
The output is a revised text passage $y$ along with an \emph{attribution report} $A$, which contains \emph{evidence snippets} $e_1, \dots, e_M$ that support the content in $y$.
Optionally, the attribution report can contain additional information such as the alignment between evidence snippets and relevant parts in $y$.

We propose to measure the quality of the revised text $y$ and attribution report $A$ along two dimensions: (1) \textbf{attribution}: how much of the revised text $y$ can be attributed to the evidence in $A$,
and (2) \textbf{preservation}: how much the revised text $y$ preserves aspects of the original text $x$.

\subsection{Measuring attribution}

Previously, \citet{Rashkin2021AIS} proposed \emph{Attributable to Identified Sources} (AIS), a human evaluation framework which considers a binary notion of attribution.
Roughly speaking, a text passage $y$ is attributable to a set $A$ of evidence if a generic hearer would affirm the statement ``According to $A$, $y$'' under the context of $y$.
A system either receives full credit (1.0) if \emph{all} content in $y$ can be attributed to $A$, and no credit (0.0) otherwise.

We propose a more fine-grained, sentence-level extension of AIS.
We ask annotators to give an AIS score for each sentence $s$ of $y$, and then report the average AIS score across all sentences:
\begin{equation}
\attrais(y, A) = \operatorname*{avg}_{s \in y} \text{AIS}(s, A).
\end{equation}
Since the AIS score is binary, this effectively measures the percentage of sentences in $y$ that are fully attributed to $A$. When judging each sentence, we also give annotators access to the surrounding sentences and other necessary context, such as the question that the text passage responded to.
We also impose the maximum number of evidence snippets in the attribution report $A$
to make it concise enough for both the annotator and downstream users.
By manually inspecting 30 examples from our benchmarks,
we found $M = 5$ snippets to be sufficient for full attribution.

During model development, we define
an automated metric, auto-AIS ($\attrauto{}$), that approximates human AIS judgments.
We utilize the natural language inference (NLI) model from \citet{Honovich2022TRUERF}, which correlates well with AIS scores.
For each sentence $s$ of $y$,
and for each evidence snippet $e$ in $A$,
let $\text{NLI}(e, s)$ be the model probability of $e$ entailing $s$. We then define
\begin{equation}
\attrauto(y, A) = \operatorname*{avg}_{s \in y} \, \max_{e \in A} \text{NLI}(e, s).
\end{equation}
To improve accuracy, we decontextualize~\cite{choi2021making} each sentence based on the entire context of $y$ %
before computing the scores.
See Appendix~\ref{sec:app-auto-eval} for implementation details.

\subsection{Measuring preservation}\label{sec:preservation}

To measure preservation, we first ask annotators to decide if the revision preserves the text's original intent (completely, somewhat, or not at all --- see Appendix~\ref{sec:app-human-eval} for exact rubrics).
Like AIS evaluation, we give annotators the necessary surrounding context.
We define the binary metric $\presintent(x, y)$ to be 1.0 if the revision completely preserves the original intent, and 0.0 otherwise.

However, even if a revision preserves intent, it may still make superfluous modifications, such as reordering words, changing textual style, or including unnecessary additional information \cite{thorne2021evidence}. Different tasks have different requirements for what should be preserved. Here, we desire a simple metric that can be readily computed for many tasks and that generally penalizes unnecessary changes.
We thus define a metric based on the character-level Levenshtein edit distance \cite{levenshtein1966binary} %
between $x$ and $y$:
\begin{equation}
\preslev(x, y) = \max\left(1 - \frac{\text{{Lev}}(x, y)}{\text{{length}}(x)}, 0\right)
\end{equation}
This metric is 1.0 if $x$ and $y$ are the same, and 0.0 if $y$ completely overwrites all parts of $x$. $\preslev$ is generally sensitive to any kind of change, but certainly does not capture all notions of preservation (e.g., preserving rhyme schemes or puns).

We want the revision to preserve the original intent while avoiding superfluous edits. To reflect this, we finally combine the two metrics as
\begin{equation}
\prescomb(x, y) = \presintent(x, y) \cdot \preslev(x, y).
\end{equation}
which is 0.0 if the revision changes the intent and equal to $\preslev(x, y)$ otherwise.
Since $\presintent$ requires human annotation,
we use $\preslev$ as an automated metric for model development.

\subsection{Discussion}

Optimizing for attribution alone cannot ensure a good revision: for example, an adversarial editor could ensure 100\% attribution by simply replacing the input $x$ with the text of any arbitrary retrieved document, which is trivially attributable to itself.
Ideally, we want to maximize both attribution and preservation, while navigating any tradoffs between the two.
In our experiments, we report both metrics, as well as their harmonic mean ($\HM{}$, analogous to how recall and precision are combined in F1).

We emphasize that this evaluation scheme does not require any ``gold'' or ``reference'' edits (unlike many prior evaluations of text revision models), which are often only available for specialized domains. This enables us to broaden the scope to a much wider range of generation tasks.

%% file: fig_overview.tex
\begin{figure}[!t]
\centering
\hspace*{-0.2cm}
\includegraphics[width=1.02\columnwidth]{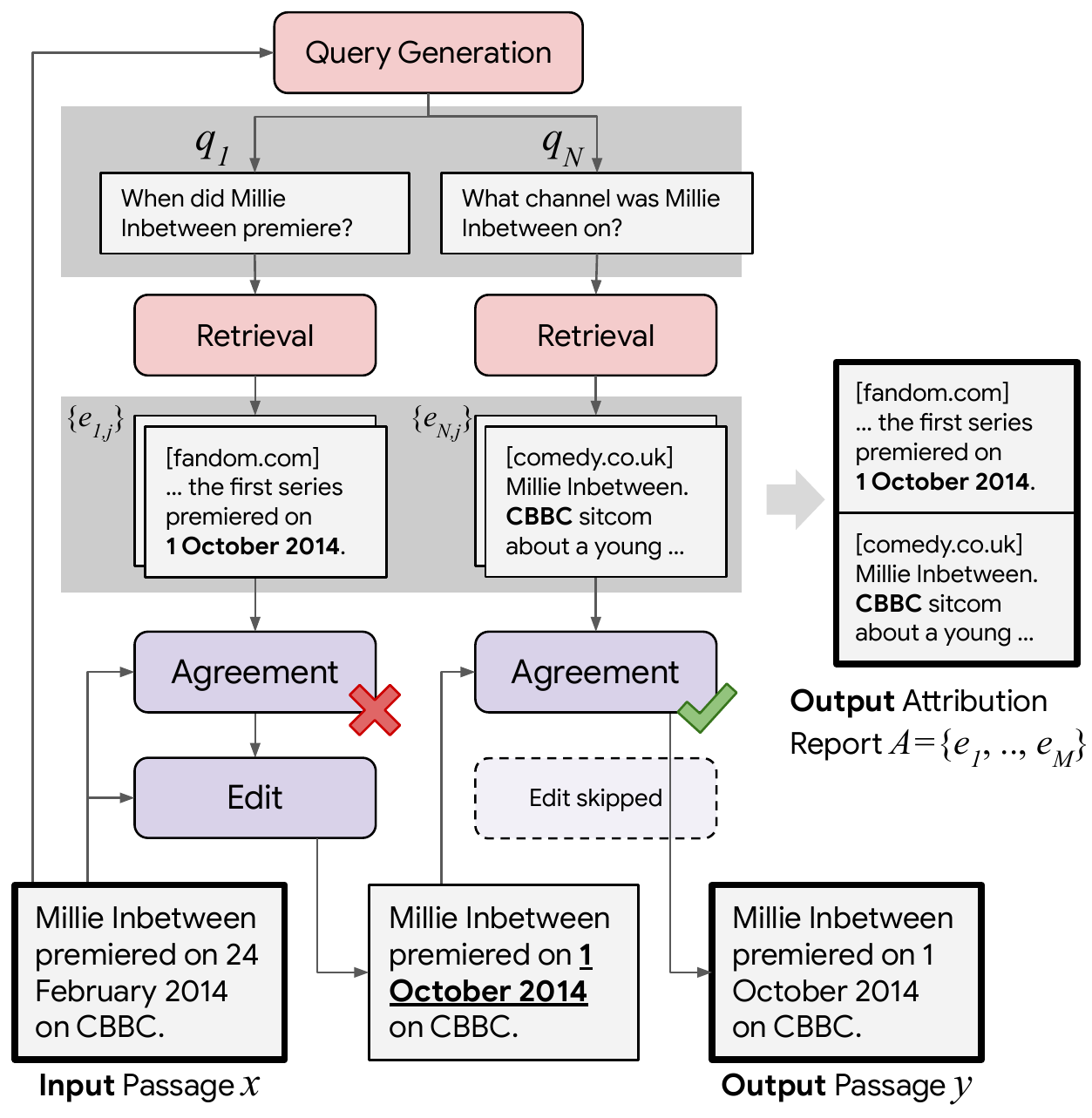}
\caption{
\label{fig:overview}
\textbf{An overview of \approach{},} which improves attribution for a text passage via \emph{Research \& Revision}. 
Given the input text passage, the \textbf{\textcolor{BrickRed}{research stage}} uses a \textit{query generator} to raise questions about different aspects of the text.
The \textit{retriever} then searches for evidence to investigate each query. The \textbf{\textcolor{DarkOrchid}{revision stage}} first runs an \textit{agreement model} to detect disagreement between the text and the evidence, then runs an \textit{edit model} to revise the text if needed.
Finally, $M$ evidence snippets are selected to form an attribution report.
}
\end{figure}

%% file: 2_approach.tex
\section{Approach}\label{section:approach}

\input{fig_example_prompts}

We now present \approachfull{} (\approach{}), a simple method for solving the \taskName{} task.
As illustrated in Figure~\ref{fig:overview}, given an input passage $x$,
the research stage first generates a set of queries $\{q_1, ..., q_N\}$, each investigating one aspect of $x$ that potentially requires attribution.
For each query $q_i$, it retrieves web documents and selects the best evidence snippets $\{\e_{i1}, \e_{i2},\dots\}$.
    The revision stage then revises the original text $x$ using the retrieval results $\{(q_1, \e_{11}), \dots\}$, yielding a revised text $y$.

Most components for \approach{} are implemented using few-shot prompting \cite{Brown2020LanguageMA}. We use PaLM \cite{Chowdhery2022PaLMSL} as our language model.
Figure~\ref{fig:example_prompts} shows some few-shot examples we use, while Appendix~\ref{section:prompting} lists the full prompts.

\subsection{Research stage}
\label{section:qgen}

\paragraph{Query generation}
We perform \emph{comprehensive question generation} (CQGen) which produces a sequence of questions covering \emph{all aspects} of the passage $x$ that need to be verified and attributed.
A similar strategy has been employed to train text-planning models \cite{narayan2022blueprint}.
A prompt with six human demonstrations %
was sufficient for PaLM to adequately learn the task.  To increase diversity and coverage, we sample from our CQGen model three times and take the union of the resulting queries.

\paragraph{Evidence retrieval}
For each query from CQGen, we use Google Search to retrieve $K=5$ web pages.
We extract candidate evidence snippets from each web page by running a sliding window of four sentences across the page, breaking at document headings.
The evidence snippets for each query are then ranked based on their relevance to the query. For this, we use an existing query-document relevance model trained following \citet{ni2021gtr}, which computes a relevance score $S_{\text{relevance}}(q, e)$ between a query $q$ and an evidence snippet $e$. We then keep the top $J=1$ evidence for each query. 
The final retrieval result is
$[
(q_1, \e_{11}),\allowbreak
\ldots,\allowbreak
(q_1, \e_{1J}),\allowbreak
\ldots,\allowbreak
(q_N, \e_{N1}),\allowbreak
\dots,\allowbreak
(q_N, \e_{NJ})]$,
where $\e_{ij}$ denotes the $j^{\text{th}}$ evidence for the $i^{\text{th}}$ query, and $N$ denotes the total number of queries from CQGen (which can be different for each input $x$).

\subsection{Revision stage}\label{sec:approach-editor}

After retrieving evidence, certain parts of $x$ may now be properly attributed, but other parts remain unattributed and should be revised.
As illustrated in Figure~\ref{fig:overview},
the revision stage initializes the output $y = x$. Then for each retrieved $(q, e) = (q_i, \e_{ij})$,
the \emph{agreement model} checks if the evidence $e$ disagrees with the current output $y$ regarding the issue in query $q$. If a disagreement is detected, the \emph{edit model} edits $y$ to agree with $e$; otherwise, it does nothing. The process continues until all retrievals are processed.

\paragraph{Agreement model}
The agreement model takes the partially edited passage $y$, a query $q$, and the evidence $e$ as input. It then decides whether both $y$ and $e$ imply the same answer to the question in $q$. This form of question-guided agreement was previously explored by \citet{honovich2021q2}.
We implement this by few-shot prompting PaLM using a chain-of-thought style prompt \cite{Wei2022COT}, where we ask the model to explicitly state the implied answers for both $y$ and $e$ before producing its judgment about their agreement. %

\paragraph{Edit model}
The edit model is run only if a disagreement is detected.
The model takes $y$, $q$ and $e$ as input, and outputs a new version of $y$ that aims to agree with $e$ while otherwise minimally altering $y$. We again use few-shot prompting and chain-of-thought, where we ask the model to first identify a particular span in $y$ that needs to be edited before generating the revised $y$. %
This helps reduce the editor's deviation from the current $y$.\footnote{The editor occasionally produces large edits that bring the new revision close to $e$ but far from the current $y$. Since this is rarely desirable, we reject edits with edit distance above 50 characters or 0.5 times the original text length.}

\subsection{Attribution report}\label{sec:attribution-report}

Finally, we select at most $M=5$ evidence snippets to form an attribution report $A$. Note that during evidence retrieval and revision, we may have encountered and used more than $M$ snippets.
Our goal is to find a subset of snippets that maximizes \emph{coverage} over the potentially attributable points in the passage, as represented by the queries ${q_1,\dots,q_N}$. We use the relevance model from Section~\ref{section:qgen} as a proxy for measuring how much an evidence $e$ covers the point raised by a query $q$. Then, we exhaustively search for $A \subseteq \{\e_{11}, \dots, \e_{NJ}\}$ of size at most $M$ that maximizes
\begin{equation}
\text{Cover}(A, q_{1:N}) := \sum_{i=1}^N \max_{e \in A} S_{\text{relevance}}(q_i, e).
\end{equation}

%% file: fig_example_prompts.tex
\newcommand{\promptin}[1]{\textcolor{blue!70!black}{{#1}}}
\newcommand{\promptout}[1]{\textcolor{red!70!black}{{\ul{#1}}}}

\begin{figure}[!t]
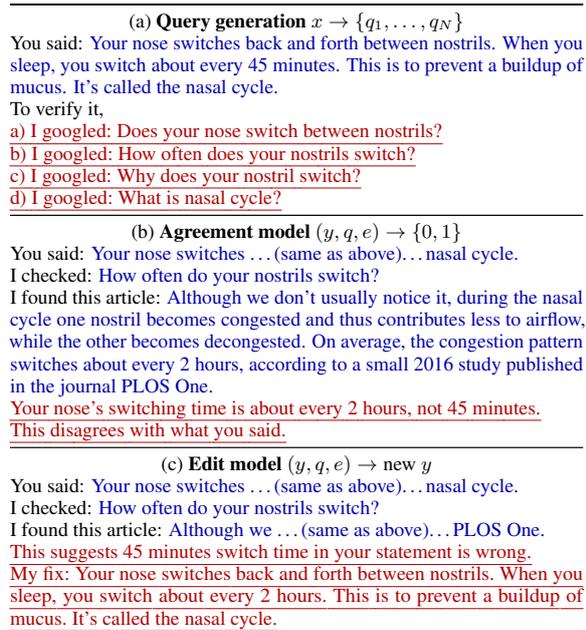

\centering\small
\scalebox{0.83}{
\begin{tabular}{@{}p{1.18\columnwidth}@{}}
\toprule
\multicolumn{1}{c}{(a) \textbf{Query generation} $x \to \{q_1, \dots, q_N\}$} \\
You said: \promptin{Your nose switches back and forth between nostrils. When you sleep, you switch about every 45 minutes.  This is to prevent a buildup of mucus. It's called the nasal cycle.} \\
To verify it, \\
\promptout{a) I googled: Does your nose switch between nostrils?} \\
\promptout{b) I googled: How often does your nostrils switch?} \\
\promptout{c) I googled: Why does your nostril switch?} \\
\promptout{d) I googled: What is nasal cycle?} \\
\midrule
\multicolumn{1}{c}{(b) \textbf{Agreement model} $(y, q, e) \to \{0, 1\}$} \\
You said: \promptin{Your nose switches \dots(same as above)\dots nasal cycle.} \\
I checked: \promptin{How often do your nostrils switch?} \\
I found this article: \promptin{Although we don't usually notice it, during the nasal cycle one nostril becomes congested and thus contributes less to airflow, while the other becomes decongested. On average, the congestion pattern switches about every 2 hours, according to a small 2016 study published in the journal PLOS One.} \\
\promptout{Your nose's switching time is about every 2 hours, not 45 minutes.} \\
\promptout{This disagrees with what you said.} \\
\midrule
\multicolumn{1}{c}{(c) \textbf{Edit model} $(y, q, e) \to \text{new }y$} \\
You said: \promptin{Your nose switches \dots(same as above)\dots nasal cycle.} \\
I checked: \promptin{How often do your nostrils switch?} \\
I found this article: \promptin{Although we \dots(same as above)\dots PLOS One.} \\
\promptout{This suggests 45 minutes switch time in your statement is wrong.} \\
\promptout{My fix: Your nose switches back and forth between nostrils. When you sleep, you switch about every 2 hours. This is to prevent a buildup of mucus. It's called the nasal cycle.} \\
\bottomrule
\end{tabular}
}
\begin{subfigure}{0pt}\phantomsubcaption\label{fig:example_prompts_cqgen}\end{subfigure}
\begin{subfigure}{0pt}\phantomsubcaption\label{fig:example_prompts_agreement}\end{subfigure}
\begin{subfigure}{0pt}\phantomsubcaption\label{fig:example_prompts_revision}\end{subfigure}
\caption{
\label{fig:example_prompts}
\textbf{Examples of few-shot examples} used to prompt the PaLM model (\promptin{blue} = input; \promptout{red} = output).
}
\end{figure}

%% file: 3_related.tex
\section{Related work}\label{section:related}

\paragraph{Fact-checking}
Our research builds upon works to identify whether a claim is supported or refuted by the given evidence \cite{thorne2018fever, wang2017liar, karadzhov2017fully, augenstein2019multifc, wadden2020fact}. In real-world scenarios such as the one which \approach{} operates in, relevant evidence may not be provided, necessitating retrieval \cite{fan2020generating,piktus2021web}.

\paragraph{Post-hoc editing for factuality}
Recent work has gone beyond checking the validity of a claim to correcting a piece of text to be factually consistent with a set of evidence via post-hoc editing \cite{shah2020automatic,thorne2021evidence,schuster2021get, Balachandran2022CorrectingDF, Cao2020FactualEC, Iso2020FactbasedTE}.
FRUIT \cite{LoganIV2022FRUIT} and PEER \cite{Schick2022PEER} both implement an editor that is fine-tuned on Wikipedia edit history with the goal of updating outdated information and collaborative writing respectively.
Evidence-based Factual Error Correction \cite[EFEC;][]{thorne2021evidence} also implements a full research-and-revise workflow trained on Wikipedia passages \cite{thorne2018fever}.
A key differentiator of \approach{} is its ability to edit the output of any generation model without being restricted by the domain, task, or the need for training data.

\paragraph{Measuring attribution}
A key part of improving attribution is being able to quantify it.
Apart from human evaluation \cite{Rashkin2021AIS},
several automated evaluation methods have been proposed.
Our work uses an entailment-based metric,
which measures whether the referenced evidence entails the output text \cite{Bohnet2022AttributedQA, Kryscinski2020EvaluatingTF,Goyal2021AnnotatingAM}.
A common alternative is to evaluate whether the output text contains the same factual information as the evidence; e.g., by checking if both yield the same answer to the same question \cite{Wang2020AskingAA}. We use this notion of attribution in \approach{}'s agreement model rather than for evaluation. 

\paragraph{Retrieval-augmented models}
Models with a retrieval component have seen successes in question answering \cite{chen2017reading,lee2019latent,nakano2021webgpt}, machine translation \cite{zhang2018guiding}, code generation \cite{hayati2018retrieval}, language modeling \cite{khandelwal2019generalization}, and other knowledge-intensive tasks \cite{lewis2020rag}.
Their retrievals are not necessarily attributions \cite{dziri2022origin,longpre2021entity}
and typically are not used to revise an existing output.
An exception is LaMDA \cite{Thoppilan2022LaMDA}, a language model for dialog that performs revision by training on human annotations.

%% file: 4_results.tex
\section{Experiments}\label{sec:experiments}

\subsection{Evaluation setups}\label{sec:eval-setup}

\input{fig_dataset_examples}

\approach{} aspires to be a general-purpose method for improving the attribution of any text generation model in any text domain.
We thus construct evaluation benchmarks by taking the task input from three diverse datasets,
and prompting different generation models to produce \emph{long-form outputs} which may contain ``hallucinations,''
as demonstrated in Figure~\ref{fig:datasets}.
These long-form outputs serve as input text passages to \approach{}.
We generate 150 development and 150 test passages for each combination of generation model and source dataset. %

\paragraph{Factoid statements}
We prompt PaLM 540B and GPT-3 text-davinci-002 to generate long-form answers to questions from the Natural Questions dev set \cite[NQ;][]{Kwiatkowski2019NaturalQA}. The resulting passages are mostly coherent but often contain factual errors. This setup examines the ability to attribute a diverse range of factoid knowledge.

\paragraph{Reasoning chains}
Language models can generate reasoning chains to answer complex questions \cite{Wei2022COT}. We use PaLM and GPT-3 to generate reasoning chains for the StrategyQA train set \cite[SQA;][]{Geva2021DidAU}.
This setup tests whether the revision model can provide better attribution for intermediate steps of reasoning, while preserving the overall reasoning process.

\paragraph{Knowledge-intensive dialogs}
We consider the conversational QA task from the QReCC dev set \cite{Anantha2021QReCCQ}.
Given the previous dialog turns, which are rounds of questions and answers ($Q_1, A_1, Q_2, A_2, \dots, Q_k$), we use LaMDA and GPT-3 to answer %
to the final question $Q_k$ conditioned on the dialog history.
The answer tends to be context-dependent, featuring pronouns and implicit references. All dialog turns are given alongside the answer as inputs to the revision model.

\subsection{Models}\label{sec:baselines}
We compare \approach{} to several systems that have a research-and-revise workflow.

\paragraph{EFEC}
We consider EFEC~\cite{thorne2021evidence} as a representative fine-tuned editor.
EFEC fine-tunes a T5-based model to revise text conditioned on multiple evidence snippets using both semi-supervised and fully-supervised approaches.
We compare against their fully-supervised approach, which performed best in their experiments.
EFEC uses a neural retrieval model~\cite{karpukhin2020dpr} to retrieve from Wikipedia; however, not all passages in our experiments are supported by Wikipedia articles.
To more fairly compare the editing capabilities of EFEC, we instead use the evidence retrieved by our research stages (CQGen and web search). %
Note that the EFEC editor conditions on multiple pieces of evidence at once, while our editor iteratively conditions on one at a time.

\paragraph{LaMDA}
LaMDA~\cite{Thoppilan2022LaMDA} generates responses in three steps: 
1) generate a ``base response'';
2) generate search queries from the base response;
3) generate a ``revised response'' conditioned on the base response and retrieved evidence.
To apply LaMDA on a given text $x$, we simply set the base response in step 1 to $x$, and then run steps 2 and 3 (we call these latter two stages ``LaMDA Research'').
LaMDA was trained as a dialog system, and always expects a dialog context where the user speaks first.
So, for non-dialog tasks, we insert an artificial user utterance as dialog history: \emph{``Tell me something interesting.''}
For the attribution report, we take all evidence documents retrieved by LaMDA during its research process.

\paragraph{\approach{}}
Our model uses few-shot prompting on PaLM 540B for query generation, the agreement model, and the edit model.
We use the same prompts for all tasks except when the context comes from a dialog,
    where we slightly modify the prompts to use the dialog context (e.g., CQGen now maps dialog context + $x$ to queries).
The query-evidence relevance model $S_\text{relevance}$ is a pretrained T5-large model \cite{raffel20t5} fine-tuned following \citet{ni2021gtr} on MS MARCO \cite{nguyen2016msmarco}.
See Appendix~\ref{app:model-details} for the few-shot prompting strategies and more modeling details.

\input{tab_main_results}

\input{fig_plot2d}

\subsection{Results}\label{sec:main-results}

For the main experiments, we report results on passages generated by PaLM and LaMDA. Results on GPT-3 passages show similar trends (Appendix~\ref{app:more-experiments}).
Table~\ref{tab:main_results} and Figure~\ref{fig:plot2d} show attribution and preservation results for each model and dataset. We also report $\HM{}$, the harmonic mean of the two metrics,
which is shown as level curves in Figure~\ref{fig:plot2d}.

\approach{} significantly improves attribution while preserving most of the original text.
In terms of $\HM{}$, \approach{} is the only method that performs robustly across all three datasets, and significantly outperforms prior methods on NQ and SQA.

We found that \approach{} is the only method that preserves the original intent of $x$ over 90\% of the time --- EFEC and LaMDA only manage to preserve the original intent 6--40\% of the time.
We also see that editing is crucial to improve attribution: if we only retrieve evidence to support the original response $x$ without editing, attribution ranges from the low 10s to mid 30s. After editing, \approach{} can increase attribution by up to 13\% absolute, while changing only 10--20\% of the text.

As noted in Section~\ref{sec:task},
one can  sacrifice preservation for higher attribution. EFEC is able to obtain strong $\HM{}$  on QReCC by making larger changes to the text in exchange for a higher attribution score. However, it occupies a very different point from \approach{} on the attribution-preservation trade-off curve, as visualized in Figure~\ref{fig:plot2d}.

\section{Analysis}\label{sec:analysis}

\subsection{Qualitative analysis}

\paragraph{Human oracle} To understand the remaining headroom in our task, we ask: \emph{what is the minimal amount of editing needed to make a text passage fully attributed?}
The answer would depend on the quality of the LM that generated the text as well as the task difficulty.
As an approximation, we manually edited 30 examples in our NQ benchmark until we judged them to be 100\% attributable. We achieved a preservation score of 88\%, which (when combined with 100\% attribution) translates to 93.6 $\HM{}$, indicating a significant headroom.

\paragraph{Analyzing the baselines}
\input{fig_compare_outputs}
As exemplified in Figure~\ref{fig:compare_outputs},
EFEC frequently attempts to summarize the entire passage into one sentence, or drops later sentences. This is likely due to EFEC's training data, which was limited to single sentences. This behavior generally increases the attribution score, because it is usually easier to make one sentence fully attributable than many sentences.
However, in datasets where the claim contains multiple sentences (NQ and SQA), such a behavior yields low preservation scores, and also results in outputs that are less informative.
We expect that EFEC could perform much better if its training data were augmented to include multiple sentences. LaMDA Research achieves similar attribution scores to \approach{}.
But as mentioned in Section~\ref{sec:baselines}, the intent and linguistic style of the output tend to deviate from the input, resulting in lower preservation scores (Figure~\ref{fig:compare_outputs}).
We emphasize that this is not a purely apples-to-apples comparison since LaMDA was not optimized for preservation.
Overall, these experiments are mainly meant to illustrate that prior models were simply not designed for the task of \taskName, rather than to mark \approach{} as the best method.

\paragraph{Analyzing \approach{}}

For the research stage,
the question generation model had comprehensive coverage:
a manual inspection of 40 examples shows > 80\% with questions that fully cover all aspects of the input text.
The retriever was strongest at researching content involving distinct entities (e.g., a movie, a major event, or a person). In contrast, we found significant headroom for better attribution of statements involving generic objects and more abstract claims (e.g. \emph{``Video games require electricity.''}--- since this is obvious to most humans, retrieved articles from the web tend to address related but different topics).
We suspect that a significant amount of attribution headroom on our benchmarks would benefit from a better  research stage.

\input{fig_interesting_examples}

For the revision stage,
\approach{} was able to revise many unattributed claims, especially those involving entities and numbers (Figures~\ref{fig:interesting_examples_entity}~and~\ref{fig:interesting_examples_number}).
It can also perform larger revisions when necessary (Figure~\ref{fig:interesting_examples_logic}).
Moreover, \approach{} abstains from editing when the claim is already well-attributed:
on NQ, among the inputs with near-perfect attribution (pre-edit $\attrais{} > 0.9$), RARR does not make an edit in 90\% of the cases.
However, the system also has several shortcomings.
Some erroneous edits arise from %
misleading irrelevant evidence (Figure~\ref{fig:interesting_examples_misleading}).
We also observed an interesting challenge when revising reasoning chains, %
where the model successfully revised an incorrect claim, but did not revise subsequent reasoning steps that depend on the earlier claim (Figure~\ref{fig:interesting_examples_reasoning_unchanged}). In this case, further editing to improve logical coherence could help.

\subsection{Ablations} \label{sec:ablations}

\input{tab_ablation_results_lower}

\paragraph{Ablating query generation}
\approach{} uses generated questions as search queries for evidence retrieval. We consider two natural alternatives: using the entire input passage as a single search query, or using each sentence as a search query.
For the former, we retrieve $J = 3$ evidence snippets to make the amount a closer match to other methods.

The results are in Table~\ref{tab:ablation_results_lower}.
Using the entire input passage as the query gives poor results,
as the retrieved evidence tends to not focus on potentially unattributed parts in the passage.
Using sentences as queries gives results closer to the full CQGen, but a closer analysis reveals two caveats.

\input{tab_wiki_ablation}

First, sentences-as-queries are more effective when such sentences ``mimic'' content on the Web,
and are less effective otherwise.
In Table~\ref{tab:wiki_ablation_results}, we test this by excluding all of Wikipedia from web search results (since many PaLM outputs for NQ have a Wikipedia style). The attribution performance of sentences-as-queries drops significantly, while CQGen is more robust.

Second, sentence-as-queries tends to retrieve passages that may encourage confirmation bias. 
Consider the example \textit{``Georgia is called the Peach State, but California actually produces the most peaches.''} Retrieval using sentences-as-queries found an article echoing that California produces the most peaches, while CQGen generated the more impartial query \textit{``Which state produces the most peaches?''} and found a newer article saying that South Carolina replaced California as the top peach producer. In this case, \approach{} using CQGen needs to sacrifice more preservation score to edit the text, leading to a lower $\HM{}$ score. This underscores that attribution alone cannot measure ``correctness'' since not all evidence is up-to-date or reliable.

\paragraph{Ablating agreement model}
We try removing the agreement model, which effectively forces the model to revise the passage based on every retrieved evidence. The results are shown in Table~\ref{tab:ablation_results_lower}. As expected, more revision leads to less preservation score and spurious changes to the text passage, as demonstrated in Figure~\ref{fig:ablate_agreement}. %

\input{fig_ablate_agreement}

\input{fig_e2e_eval}

\paragraph{Impact on downstream task performance}\label{sec:e2e_eval}
We have measured preservation using the metric defined in Section~\ref{sec:preservation}. However, another measure of preservation is whether the revised text can still be used to perform the task that it was originally generated for. 
Following EFEC,
we quantitatively evaluate this on short answer tasks NQ and SQA, and we summarize the result in Figure~\ref{fig:e2e}.

For NQ, each original text $x$ is a long-form response to a factoid question. To determine whether the revised text $y$ still serves this purpose, we feed the factoid question and $y$ back into PaLM and prompt it to extract a \emph{short} answer from $y$.
We find that \approach{} not only preserves the short answer accuracy but actually improves it by roughly 5\%.

For SQA, each original text is a reasoning chain that helps to answer a yes/no question. 
We feed the SQA question and $y$ back into PaLM and prompt it to output a yes/no answer, and evaluate answer accuracy. Here, we find that increasing attribution comes at a slight cost in downstream task performance: answer accuracy drops modestly for \emph{all} revision models (up to 2.6\%). 
We suspect that this may be due to noisy retrievals, which sometimes provide misleading evidence (exemplified in Figure~\ref{fig:interesting_examples_misleading}).
Furthermore, even though revisions can address factoid errors in the passage (e.g., \emph{``Homer Simpson has 5 fingers''} from Figure~\ref{fig:interesting_examples_reasoning_unchanged}), \approach{} currently does not try to modify subsequent reasoning steps which may no longer be logically entailed (e.g., \emph{``He only needs one hand to count to 5''}).

%% file: fig_dataset_examples.tex
\begin{figure}[!t]
\centering\footnotesize
\adjustbox{max width=\columnwidth}{%
\begin{tabular}{@{}p{1.15\columnwidth}@{}}
\toprule
\multicolumn{1}{c}{\textbf{PaLM outputs on NQ (factoid statements)}} \\
Millie Inbetween is a British comedy television series. It premiered on 24 February 2014 on BBC One. The first series was produced by John Yorke and Phil Clymer. \\ \midrule
\multicolumn{1}{c}{\textbf{PaLM outputs on SQA (reasoning chains)}} \\
The highest point of Mount Wycheproof is 70 metres. Edmund Hillary climbed Mount Everest, which is 8,848 metres. So Mount Wycheproof would be a breeze for Edmund Hillary. \\ \midrule
\multicolumn{1}{c}{\textbf{LaMDA outputs on QReCC (knowledge-intensive dialogs)}} \\
\begin{tabular}{@{}ll@{}}
When was Welsh social reformer Robert Owen born?
& \multirow{4}{*}{
$\left.\vphantom{\rule{0pt}{5ex}}\right\}$ context
} \\
Robert Owen was born on 14 May 1771 \\
\dots \\
Did he have another job? \\
\end{tabular} \\
In 1810 he moved to Manchester and established a draper's shop. \\
\bottomrule
\end{tabular}
}
\caption{
\label{fig:datasets}
\textbf{Examples of input passages.} For QReCC, prior dialog turns are also given as the context.
}
\end{figure}

%% file: tab_main_results.tex
\newcommand{\mainTableDatasetName}[1]{\multicolumn{9}{c}{\textbf{#1}}}

\begin{table}[!t]
\setlength{\tabcolsep}{4pt}
\centering
\adjustbox{max width=\columnwidth}{%
\begin{tabular}{
@{}l
r@{ $\to$ }r
r@{ $\to$ }r
rrrr
@{}}
\toprule
 & \multicolumn{4}{c}{Attribution} & \multicolumn{3}{c}{Preservation}
 \\
 \cmidrule(lr){2-5} \cmidrule(lr){6-8}
 Model
 & \multicolumn{2}{c}{auto-AIS} & \multicolumn{2}{c}{AIS} & \MN{intent} & \MN{Lev} & \MN{comb} & $\HM{}$
 \\
\midrule
\mainTableDatasetName{PaLM outputs on NQ} \\
EFEC
& 45.6 & 64.3 & 35.4 & 48.3 & 16.0 & 39.1 & 10.4 & 17.1 \\
LaMDA
& 39.5 & 49.9 & 18.3 & 30.4 & 26.0 & 39.6 & 21.1 & 24.9 \\
\approach{}
& 45.6 & 54.9 & 35.4 & 43.4 & 90.0 & 89.6 & 83.1 & \textbf{57.0} \\ \midrule
\mainTableDatasetName{PaLM outputs on SQA} \\
EFEC
& 37.8 & 58.6 & 24.5 & 51.7 & 6.0 & 31.0 & 3.8 & 7.1 \\
LaMDA
& 32.7 & 43.2 & 15.8 & 27.0 & 40.0 & 46.4 & 33.7 & 30.0 \\
\approach{}
& 37.6 & 45.1 & 24.5 & 31.5 & 92.6 & 89.9 & 84.6 & \textbf{45.9} \\ \midrule
\mainTableDatasetName{LaMDA outputs on QReCC} \\
EFEC
& 19.1 & 47.4 & 13.2 & 48.7 & 39.7 & 39.4 & 23.7 & 31.9 \\
LaMDA
& 16.4 & 36.2 & 16.0 & 27.1 & 21.3 & 24.8 & 12.0 & 16.6 \\
\approach{}
& 18.8 & 29.4 & 13.2 & 28.3 & 95.6 & 80.2 & 78.1 & \textbf{41.5} \\
\bottomrule
\end{tabular}
}
\caption{\textbf{Evaluation results.} For attribution, we report the AIS scores of the texts both before and after editing (before $\rightarrow$ after). For preservation, we report intent preservation $\presintent$, Levenshtein similarity $\preslev$, and the combined $\prescomb$. We summarize $\attrais$ and $\prescomb$ using their harmonic mean ($\HM{}$). %
}
\label{tab:main_results}
\end{table}

%% file: fig_plot2d.tex
\begin{figure}[!t]
\centering
\includegraphics[width=\linewidth]{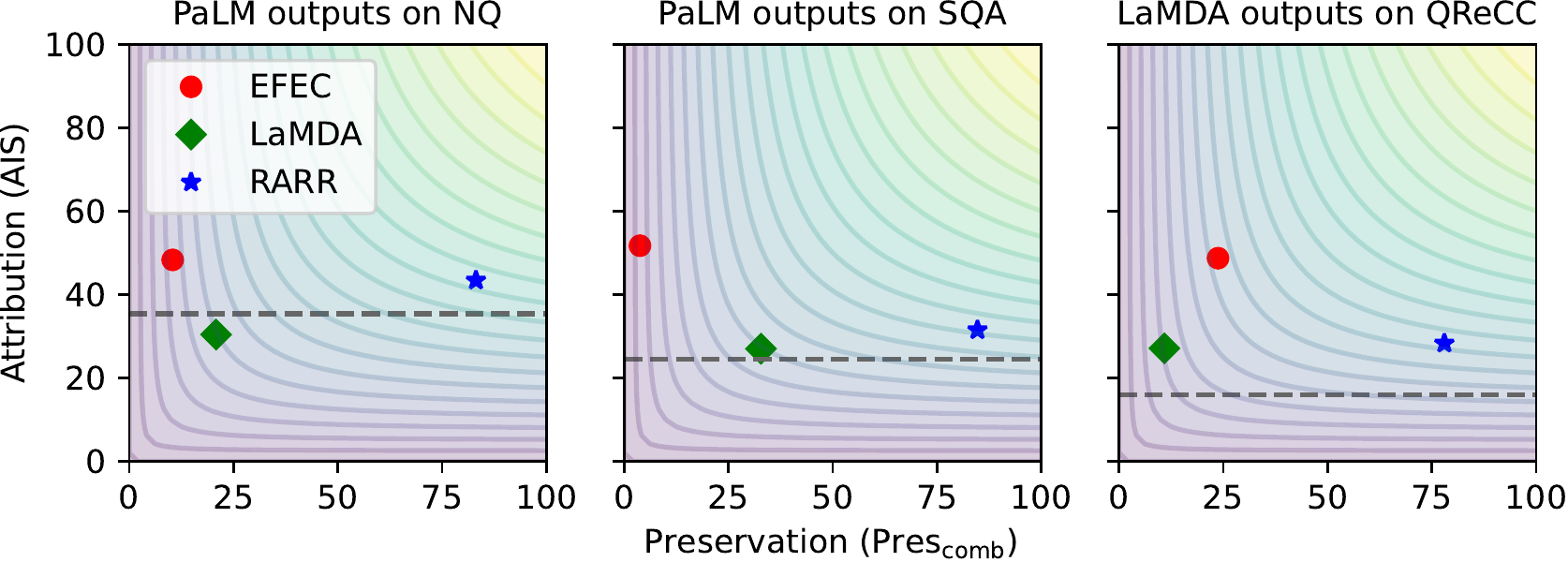}
\caption{
\label{fig:plot2d}
\textbf{Attribution and preservation scores.}
Dashed lines indicate the highest attribution score obtained by any of the models \emph{before} editing: points above the line have better attribution after revision.
The contours are $\HM{}$ level curves: points along a contour have equivalent $\HM{}$. Different models make very different trade-offs between attribution and preservation. Only \approach{} has a robust $\HM{}$ across all tasks.
}
\end{figure}

%% file: fig_compare_outputs.tex
\begin{figure}[!t]
\small
\centering
\adjustbox{max width=\columnwidth}{%
\begin{tabular}{
@{\!\!}r@{\hspace{3pt}}m{0.5\columnwidth}l@{}
}
\toprule

$x$: & \cmpR{Justice Ashok Kumar Mathur headed the 7th central pay
   commission in India. It was created in 2014 and submitted its
   report in {\color{red}\textbf{2016}}.} \\
& \textbf{Attribution:} 50\% & \textbf{Preservation:} 100\% \\
\midrule
EFEC: & \cmpR{The 7th central pay commission in India was created in 2014.} \\
& \textbf{Attribution:} 100\% & \textbf{Preservation:} 0\% \\
\midrule
LaMDA: & \cmpR{
I heard the 7th CPC made recommendations for
  increasing the minimum salary pay from Rs 7066 to 18k per month for
  new central government employees.} \\
& \textbf{Attribution:} 0\% & \textbf{Preservation:} 0\% \\
\midrule
\approach{}: & \cmpR{
Justice Ashok Kumar Mathur headed the 7th
central pay commission in India. It was created in 2014 and
submitted its report in {\color{blue}\textbf{2015}}.
} \\
& \textbf{Attribution:} 100\% & \textbf{Preservation:} 100\% \\
\midrule
evidence: & \cmpR{
The 7th Central Pay Commission (Chair: Justice A. K. Mathur)
\textbf{submitted its report on November 19, 2015}. The Commission had been
\textbf{appointed in February 2014}, to look at remuneration for central
government employees. \dots %
}
\\
\bottomrule
\end{tabular}
}
\caption{
\label{fig:compare_outputs}
\textbf{Example model outputs and human judgment of their attribution and preservation scores.}
EFEC reduces the passage $x$ into a single sentence. LaMDA changes the writing style.
\approach{} preserves the structure of the input passage. We show one evidence retrieved by \approach{} to help explain the example.
}
\end{figure}

%% file: fig_interesting_examples.tex
\begin{figure}[!t]
\centering\small %
\adjustbox{max width=\columnwidth}{%
\begin{tabular}{@{}r@{\hspace{4pt}}p{1.25\columnwidth}@{}}
\toprule
\multicolumn{2}{c}{(a) \textbf{Correctly revising an entity}} \\
$y$: & If She Knew What She Wants was written by {\color{red}\textbf{Henry Roth}}. \\
$e$: & [en.wikipedia.org] ``If She Knew What She Wants'' is a song written by American singer-songwriter \textbf{Jules Shear} and introduced on \dots \\ %
$y'$: & If She Knew What She Wants was written by {\color{blue}\textbf{Jules Shear}}. \\
\midrule
\multicolumn{2}{c}{(b) \textbf{Correctly revising a number}} \\
$y$: & God Save the Queen became the British national anthem in {\color{red}\textbf{1745}}. \dots \\
$e$: & [www.britannica.com] The oldest national anthem is Great Britain's ``God Save the Queen,'' which was described as a national anthem in \textbf{1825}, \dots \\ %
$y'$: & God Save the Queen became the British national anthem in {\color{blue}\textbf{1825}}. \dots \\
& \emph{(The year 1745 was when the song was first performed.)} \\
\midrule
\multicolumn{2}{c}{(c) \textbf{Performing a necessary larger revision}} \\
$y$: & ``It's My Party'' is a song written and composed by American singer-songwriter and producer {\color{red}\textbf{Walter Gold}}. The song was published in {\color{red}\textbf{1962}} and first recorded by American singer-songwriter, {\color{red}\textbf{producer and arranger}} Lesley Gore. \\
$e$: & [secondhandsongs.com] Song: It's My Party written by \textbf{John Gluck, Wally Gold, Seymour Gottlieb, Herbert Weiner}\dots \\
$y'$: & ``It's My Party'' is a song written and composed by {\color{blue}\textbf{John Gluck, Wally Gold, Seymour Gottlieb and Herb Weiner}}. The song was published in {\color{blue}\textbf{1963}} and first recorded by American {\color{blue}\textbf{singer}} Lesley Gore. \\

\midrule
\multicolumn{2}{c}{(d) \textbf{Confused by misleading evidence}} \\
$y$: & In the Heat of the Night is a police procedural crime drama TV series produced by MGM Television. It aired on {\color{blue}\textbf{NBC}} from September 8, 1988 to May 3, 1992. \dots \\
$e$: & [metv.com] Watch In the Heat of the Night on \textbf{MeTV}. \\
$y'$: & In the Heat of the Night is a police procedural crime drama TV series produced by MGM Television. It aired on {\color{red}\textbf{MeTV}} from September 8, 1988 to May 3, 1992. \dots \\
& \emph{(The show did actually air on NBC in the 1980s, but the model fails to realize that MeTV is a re-run much later in time.)} \\
\midrule
\multicolumn{2}{c}{(e) \textbf{Improved attribution, but reasoning doesn't adapt to new info}} \\
& \textit{(Q: Does Homer Simpson need two hands worth of fingers to count to 5?)} \\
$y$: & Homer Simpson has {\color{red}\textbf{5}} fingers and 2 hands. Therefore, he does not need two hands worth of fingers to count to 5. He only need one hand to count to 5. \\
$e$: & [refinery29.com] Other iconic animated characters, such as Homer Simpson and his brood, \dots only have \textbf{four fingers}, too.\\
$y'$: & Homer Simpson has {\color{blue}\textbf{4}} fingers and 2 hands. Therefore, he does not need two hands worth of fingers to count to 5. He only need one hand to count to 5. \\
& \emph{(Having only  4 fingers per hand, he does need two hands to count to 5.)} \\

\bottomrule
\end{tabular}
}
\begin{subfigure}{0pt}\phantomsubcaption\label{fig:interesting_examples_entity}\end{subfigure}
\begin{subfigure}{0pt}\phantomsubcaption\label{fig:interesting_examples_number}\end{subfigure}
\begin{subfigure}{0pt}\phantomsubcaption\label{fig:interesting_examples_logic}\end{subfigure}
\begin{subfigure}{0pt}\phantomsubcaption\label{fig:interesting_examples_misleading}\end{subfigure}
\begin{subfigure}{0pt}\phantomsubcaption\label{fig:interesting_examples_reasoning_unchanged}\end{subfigure}
\caption{
\label{fig:interesting_examples}
\textbf{Example revisions  from \approach{}, both good and bad.}
$y$ = partially edited passage; $e$ = evidence; $y'$ = passage after editing with $e$.
}
\end{figure}

%% file: tab_ablation_results_lower.tex
\begin{table*}[!t]
\centering\small\setlength{\tabcolsep}{5pt}
\adjustbox{max width=\textwidth}{%
\begin{tabular}{
@{}l
r@{ $\to$ }rcc
r@{ $\to$ }rcc
r@{ $\to$ }rcc
@{}}
\toprule
 & \multicolumn{4}{c}{PaLM outputs on NQ}
 & \multicolumn{4}{c}{PaLM outputs on SQA}
 & \multicolumn{4}{c}{LaMDA outputs on QReCC}
 \\
 \cmidrule(lr){2-5} \cmidrule(lr){6-9} \cmidrule(lr){10-13}
 Model
 & \multicolumn{2}{c}{$\attrauto{}$} & $\preslev{}$ & \multicolumn{1}{c}{$\HM{}$}
 & \multicolumn{2}{c}{$\attrauto{}$} & $\preslev{}$ & \multicolumn{1}{c}{$\HM{}$}
 & \multicolumn{2}{c}{$\attrauto{}$} & $\preslev{}$ & \multicolumn{1}{c@{}}{$\HM{}$}
 \\
\midrule
Full \approach{}
& 45.6 & 54.9 & 89.6 & \textbf{68.1}
& 37.6 & 45.1 & 89.9 & 60.0
& 18.8 & 29.4 & 80.2 & \textbf{43.1}
\\ %
no agreement model
& 45.6 & 50.6 & 82.6 & 62.8
& 37.8 & 46.9 & 83.4 & 60.0
& 18.8 & 28.8 & 72.0 & 41.2
\\ %
query = input
& 45.4 & 47.2 & 98.4 & 63.8
& 39.4 & 30.3 & 98.8 & 46.4
& 19.7 & 20.6 & 96.3 & 34.0
\\
query = sentence
& 49.1 & 52.1 & 97.0 & 67.8
& 43.7 & 44.3 & 98.8 & \textbf{61.2}
& 19.0 & 19.6 & 97.0 & 32.6
\\
\bottomrule
\end{tabular}
}
\caption{\textbf{Ablation results.} We report the automatic metrics: $\attrauto$, $\preslev$, and harmonic mean between the two ($\HM{}$). We show auto-AIS scores both before and after editing (before $\rightarrow$ edit), with respect to the attribution report $A$ produced by the model.
Even though sentence-as-queries may achieve similar $\HM{}$ as \approach{}, it is less robust to corpus shifts and tends to retrieve passages that may encourage confirmation bias.
}
\label{tab:ablation_results_lower}
\end{table*}

%% file: tab_wiki_ablation.tex
\begin{table}[!t]
\centering\small
\adjustbox{max width=\columnwidth}{%
\begin{tabular}{@{}lcccc@{}}
\toprule
 & \multicolumn{2}{c}{NQ $\HM{}$}
 & \multicolumn{2}{c}{SQA $\HM{}$}
 \\ \cmidrule(lr){2-3} \cmidrule(lr){4-5}
 Model
 & \multicolumn{1}{c}{orig} &  \multicolumn{1}{c}{no wiki}
 & \multicolumn{1}{c}{orig} &  \multicolumn{1}{c@{}}{no wiki} \\
\midrule
Full RARR
& \textbf{68.1} & \textbf{64.3}  & 60.0 & \textbf{57.6} 
\\
query = sentence
& 67.8 & 60.3 & \textbf{61.2} & 56.7
\\
\bottomrule
\end{tabular}
}
\caption{
\textbf{The impact of excluding Wikipedia from the retrieval corpus.} %
CQGen (full \approach{}) is more robust to Wikipedia's absence, while using sentences-as-queries suffers a bigger drop in performance.}
\label{tab:wiki_ablation_results}
\end{table}

%% file: fig_ablate_agreement.tex
\begin{figure}[!t]
\centering\footnotesize
\adjustbox{max width=\columnwidth}{
\begin{tabular}{@{}r@{\hspace{4pt}}p{1.2\columnwidth}@{}}
\toprule
$x$: & The Crown-of-thorns starfish is native to the Great Barrier Reef\dots The starfish was introduced to the Great-Barrier-Reef by \textbf{ocean currents}. \\
$e$: & [invasivespeciesinfo.gov] \textbf{Ballast water} is one of the major pathways for the introduction of nonindigenous marine species\dots \\
$y$: & The Crown-of-thorns starfish is native to the Great Barrier Reef\dots The starfish was introduced to the Great-Barrier-Reef by {\color{red}\textbf{ballast water}}. \\
\bottomrule
\end{tabular}
}
\caption{
\label{fig:ablate_agreement}
\textbf{Disabling the agreement model leads to over-edits.} Here, the evidence $e$ does not explicitly disagree with $x$, but without an agreement model to detect this, the edit model makes an unsupported change.
}
\end{figure}

%% file: fig_e2e_eval.tex
\begin{figure}[!t]
\centering
\hspace*{-0.2cm}
\includegraphics[width=0.98\columnwidth]{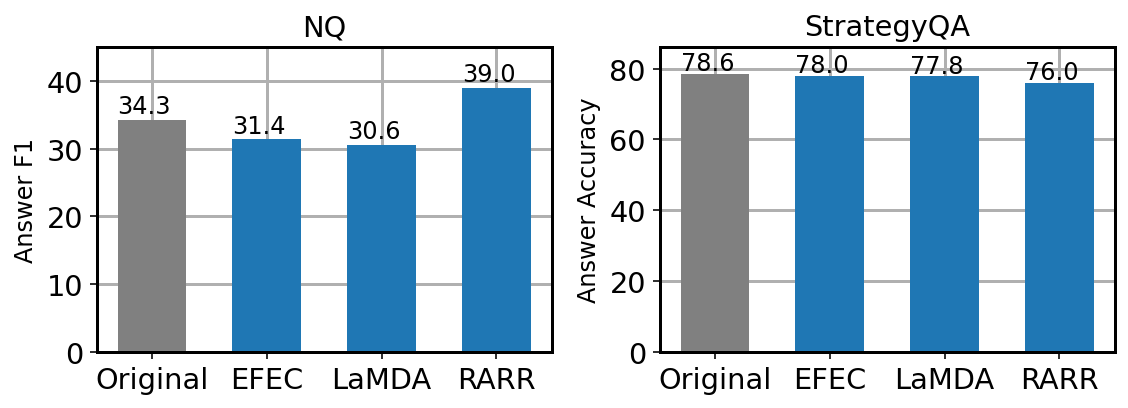}
\caption{
\label{fig:e2e}
\textbf{Downstream task performance} on NQ and SQA. \approach{}'s revisions lead to better answer accuracy on NQ. No models improved answer accuracy on SQA. %
}
\end{figure}

%% file: 5_discussion.tex
\section{Conclusion}
Language models have developed increasingly good ``procedural'' knowledge of what should be discussed and how it should be presented, but often struggle to memorize ``factoid'' knowledge and produce unsubstantiated claims.
We proposed \approach{}, a framework for revising such claims to make them attributable to the researched evidence.
From experiments on text passages generated by different models on various domains, we showed that \approach{} can revise the passages to improve attribution while preserving other desirable properties such as writing style or structure.
Furthermore, \approach{} sits on top of existing generation models without needing to re-design or re-train LMs.

Major headroom still remains, as discussed in Section~\ref{sec:analysis} and the Limitations section. We hope our analysis of \approach{} would help with developing new approaches for integrating attribution to LMs.

\section{Limitations}

\paragraph{Limitations of our task definition}
Depending on the application, attribution and preservation may not deserve equal weight. For instance, if there are multiple acceptable options for the output, such as in a dialog system, we might trade-off preservation for attribution, similar to how LaMDA behaves in our experiments.

Our evaluation metrics also do not measure all aspects of attribution. For instance, some sentences are self-evident and do not require attribution (e.g., \emph{``I agree.''}) but would be penalized in our evaluation.
It is also necessary to note that linguistic assertions have varying scope: for example, there is a difference between \emph{``Frozen is a scary movie''} and \emph{``I got scared watching Frozen''} --- while expressing a similar sentiment, the former makes a more general statement that many would disagree with, while the latter is scoped to the speaker's own experience. In some applications, one could even argue that the latter case does not require attribution, since the speaker is their own source-of-truth. In addition to varying scope, utterances can also make assertions with varying levels of directness. For example, according to standard linguistics, \emph{``John ate some of the cookies''} yields the implicature that John did not eat \emph{all} of the cookies, even though it is not logically entailed. This raises the question of which implicatures or implied assertions should be detected and attributed, which should be explored in future work. For more nuances, we refer to \citet{Rashkin2021AIS}.

For preservation, we wish to explore other properties that should be preserved, such as discourse or logical coherence. Additionally, if the input text passage is completely misguided or flawed, it can be difficult to revise the text without significant changes, which would be heavily penalized by the current metrics.

\paragraph{Limitations of our model}
While we aspire to improve attribution for arbitrary text, it is clear that \approach{} is not yet fully general. For example, the current implementation of \approach{} would not be well-prepared to edit poetry (where preserving rhyme matters) or long documents, primarily because we do not provide examples of such inputs in our few-shot LLM prompts. However, we do believe that future developers may be able to quickly adapt \approach{} to such tasks by simply changing the prompts. Second, \approach{} tends to preserve rather than delete claims that it cannot attribute. Some of these claims genuinely do not require attribution, but others are hallucination and should be removed. Judging whether a claim requires attribution can be subjective and challenging. Finally, our model is computationally costly, since it is based on prompting a large language model. One potential solution is to leverage recent synthetic data generation recipes to train a smaller model \cite{lee2021neural,Schick2022PEER}.

\section{Ethical considerations}

\paragraph{Partial attribution}
When \approach{} is not 100\% successful in making text consistent with retrieved evidence, the revised text will be partially attributed.
One could identify unattributed parts using either the automated attribution score ($\attrais{}$) or the relevance scores used to generate the attribution report (Section~\ref{sec:attribution-report}).
Such information should be presented to avoid misleading readers into thinking that the entire revision is attributed.

\paragraph{Evidence trustworthiness}
\approach{} seeks to improve attribution for the output of any generative model. However, even if \approach{} can attribute content to a particular source, the user must still consider whether the source itself is trustworthy. Even for sources that are traditionally considered ``authoritative'' (such as an encyclopedia), there may still be factual inaccuracies or biases. This work does not address the question of whether a source is trustworthy, or the related topic of misinformation.
While we do not provide a means for judging trustworthiness, the design of \approach{} does allow for the research stage to restrict its search over a user-specified corpus, based on what the user deems trustworthy.

\paragraph{Conflicting evidence}
There is also the possibility that some content may be simultaneously supported by certain sources, while contradicted by others. This can easily occur for content involving subjective or imprecise claims. The current implementation and evaluation for \approach{} does not explicitly address this issue --- we adopted a ``permissive'' definition of attribution, where we consider content to be attributed if there exists any source that supports it. For some applications, a more restrictive definition that requires both existence of supporting sources and absence of contradicting sources would be needed.

%% file: 6_contributions.tex
\section*{Acknowledgments}

We wish to thank Raphael Hoffmann, Slav Petrov, Dipanjan Das, Michael Collins, Iftekhar Naim, Kristina Toutanova, William Cohen, Sundeep Tirumalareddy, Samer Hassan, Quoc Le and Heng-Tze Cheng for their research mentorship, feedback and support. We are grateful to Hao Zhou and Petr Pilar for helping us experiment with LaMDA and motivating our dialog experiments. We also wish to thank Tal Schuster for pointing us to relevant work in the fact checking literature, and helping us reproduce it. We thank Vitaly Nikolaev, David Reitter and Roee Aharoni for helping us use AIS and auto-AIS. We also wish to thank Jianmo Ni and Honglei Zhuang for developing the query-evidence relevance model we use, Daniel Andor for developing the sentence decontextualization model we use, and Ran Tian for the initial prototype of CQGen. Finally, we thank Kathy Meier-Hellstern, Philip Parham and Diane Korngiebel for their thoughtful feedback on ethical considerations.

\section*{Contributions}

\textbf{Luyu Gao}: Designed \approach{}'s few-shot prompting strategies and implemented the first PaLM-based prototype. Analyzed results, and advised on the design of human and automatic evaluation.

\textbf{Zhuyun Dai}: Proposed the evaluation setup of editing long-form generations from PaLM/LaMDA on various QA datasets. Hosted and mentored Luyu Gao (student researcher) in prototyping \approach{}. Implemented the final models, designed overall experiments, and obtained main results and ablations (together with Ice Pasupat). Contributed many parts of the writing.

\textbf{Ice Pasupat}: Implemented the final models, designed overall experiments, and obtained main results and ablations (together with Zhuyun Dai). Automated experimental infrastructure, conducted error analyses, and oversaw many parts of the paper writing.

\textbf{Anthony Chen}: Developed the automatic evaluation for attribution and preservation and worked with Arun Chaganty to design human evaluation. Developed the open-source implementation (GPT-3 \approach{}), made improvements to prompts, and helped with writing.

\textbf{Arun Chaganty}: Led and implemented all human evaluation. Proposed the two-dimensional attribution + preservation metric (together with Kelvin Guu). Advised on model design and contributed many parts of the writing.

\textbf{Yicheng Fan}: Worked with Kelvin Guu to develop the first prototype of \approach{}. Proposed multiple retrieval strategies and implemented the EFEC baseline.

\textbf{Vincent Zhao}: Co-hosted and mentored Luyu Gao (student researcher) in prototyping \approach{}. Enabled bulk inference for PaLM. Proposed the downstream task evaluation.

\textbf{Ni Lao}: Research mentorship, advising and contributed many parts of the writing.

\textbf{Hongrae Lee}: Research mentorship and advising. Helped integrate \approach{} with Google Search and evaluate LaMDA.

\textbf{Da-Cheng Juan}: Research mentorship and early design discussions.

\textbf{Kelvin Guu}: Proposed the original research-and-revise concept, implemented the first prototype, initiated the project and involved all collaborators. Implemented baselines (together with Yicheng Fan). Research mentorship, oversaw project coordination and paper writing.

%% file: 7_appendix.tex
\section{Additional experiments and analysis} \label{app:more-experiments}

\paragraph{Model variance}
The main experiments in Section~\ref{sec:experiments} are based on a single run.
We ran automated evaluation on 3 random runs of \approach{}, using PaLM outputs on NQ as input passages.
The standard deviations of $\attrauto{}$, $\preslev{}$, and $\HM{}$ are 1.2, 0.5, and 1.0 respectively.

\paragraph{Impact of the retriever choice}
We tried using Microsoft Bing in place of Google Search, with near identical results (< 1\% difference).

\input{tab_ablation_results_upper}

\paragraph{Impact of model scale}
Many components in \approach{} work by few-shot prompting PaLM, a large 540B  parameter LM. To assess the benefit of LM scaling, we replaced PaLM 540B with a smaller 62B parameter PaLM. As shown in Table~\ref{tab:ablation_results_upper}, we found that 540B outperforms 62B by a large margin, suggesting that \approach{} could potentially further improve with even more scaling. We also experimented with keeping the editor stage at 540B while shrinking the query generation stage to 64B --- this yielded a relatively small performance drop, suggesting that model scaling is more important for the editor.

\paragraph{Impact of model type}
Few-shot prompting has proven to be effective for many recent large language models. We try replacing the query generation model, agreement model, and edit model with GPT-3 text-davinci-003. The few-shot prompts were slightly tuned to fit the GPT-3 model.
Table~\ref{tab:ablation_results_upper} shows the results, which are slightly better than \approach{} implemented with PaLM 540B on all three datasets.
We will release this open-source version of RARR that uses GPT-3 as the backbone.

\input{tab_gpt3_passages_results}

\paragraph{Results on GPT-3 passages}
Table~\ref{tab:gpt3_passages_results} shows automated evaluation results on passages generated by GPT-3.
The results follow the same trend as the results on PaLM and LaMDA passages.

\paragraph{Challenging domains}
We report results on tasks where attribution was particularly hard, and significant future work is needed.

\input{tab_eli5_and_summeval}

We considered news article summaries produced by summarization models from SummEval \citep{fabbri21summeval} (e.g., \textit{``John Doe was left homeless when the storms hit Staten Island, New York \dots''}). Results are shown in Table~\ref{tab:eli5_and_summeval}. First, we note that the before-edit auto-AIS scores for all models are low. These news article summaries are often about less widely known people and events, which is challenging for retrievers, leading to low attribution. For example, our query generator may ask \emph{``where does John Doe live''} but get results for a different John Doe. EFEC and LaMDA also face this issue, but instead trade preservation for attribution and rewrite the text to a different topic. 
This result suggests that using web search with standard question generation methods may fail to capture important context from the input, and is not sufficient for the attribution task.

We also considered long-form explanations generated by PaLM for the ELI5 dataset~\citep{fan19eli5} (Table~\ref{tab:eli5_and_summeval}). ELI5 was collected from online forums, so many answers tend to have subjective opinions instead of specific entities and facts (e.g., \textit{``How do our brains interpret scary music? To me, scary music often sounds a little bit like a person \dots''}), and are thus difficult to attribute. Sometimes the whole output is based on a false premise and needs to be completely rewritten, in which case \approach{} cannot satisfactorily edit due to our revision threshold (Section~\ref{sec:approach-editor}). %

\input{tab_mmlu}

Finally, we considered technical explanations to questions from the MMLU dataset \cite{hendryckstest2021} which covers diverse subjects from social science, humanities, STEM, and others.\footnote{MMLU has questions from 57 subjects; we took 10 random question from each topic and generated answer explanations by prompting PALM 540B.} An example input looks like \textit{ ``Every time you remove an edge from a complete graph, you divide it into two connected components. So, a complete graph with 13 vertices must have 12 connected components.''} Results are shown in Table~\ref{tab:mmlu}. \approach{} improves attribution of the explanations on all four categories of MMLU, although the increases are relatively small. We also found that \approach{}'s performance is low on examples with mathematical reasoning, as these are beyond the capability of the edit model with our current prompt.

\section{Details on automated evaluation}\label{sec:app-auto-eval}
\paragraph{Sentence splitting} When computing the attribution score, we use spaCy \verb|en_core_web_sm| v3.0.0a1 to segment the text passage into sentences. (More recent models gave similar results.) While each sentence may contain multiple claims that could be attributed independently, there is currently no linguistic consensus on what constitutes a claim. Instead of depending on a particular definition of claims, we use sentences as claims for simplicity and reproducibility. The same segmentation is also used for human evaluation.

\paragraph{Decontextualization} We decontextualize each sentence in the text passage before computing the attribution score. We use the model from \citet{choi2021making}, which is a T5 model fine-tuned to map the input ``\texttt{[HEAD] [SEP] }\textit{context and passage}\texttt{ [start] }\textit{sentence}\texttt{ [end]}'' to the output ``\texttt{[OPCODE] }\textit{decontextualized sentence}'', where the \texttt{OPCODE} can be ``\texttt{done}'' (success), ``\texttt{un}'' (unnecessary), or ``\texttt{imp}'' (impossible). We feed the passage's context (questions for NQ and SQA; dialog context for QRECC) along with the passage itself to the input. We use beam search with beam size 8 and discard any result whose number of tokens differ by more than 4.

\paragraph{NLI model} We obtained a newer version of the end-to-end NLI model from the authors of \citet{Honovich2022TRUERF}, which was trained on MNLI, SNLI, FEVER, PAWS, SciTail and VitaminC \cite{williams2017broad,bowman2015large,thorne2018fever,zhang2019paws,khot2018scitail,schuster2021get}.
The model is a T5 model fine-tuned to map the input ``\texttt{premise: }\textit{evidence}\texttt{ hypothesis:   }\textit{claim sentence}'' to either ``\texttt{1}'' (entailed) or ``\texttt{0}'' (not entailed). As suggested by the authors, we use the probability of producing ``\texttt{1}'' as the entailment score.

\input{fig_human_vs_auto_eval}

\paragraph{Comparing human and automated evaluation}
We conducted correlation studies between human and automatic metrics and found strong Pearson correlation (attribution = 0.74; preservation = 0.62).
We visualize the correlation between human and automated attribution scores on NQ and SQA in Figure~\ref{fig:human-vs-auto-eval}. We found that the AIS scores from human correlate well with auto-AIS scores, with some bias for non-attributed sentences to be judged as attributed by auto-AIS.

\input{7a_human_eval}

\section{Details on the model}
\label{app:model-details}

\paragraph{Few-shot prompting with LLMs} \label{section:prompting}

We implement many sub-tasks within \approach{} using \emph{few-shot prompting} of LLMs (also known as \emph{in-context learning}~\cite{Brown2020LanguageMA}) as follows:
\begin{enumerate} \setcounter{enumi}{0}
\item
For each sub-task, we manually author a small number of training examples: \texttt{$(\text{input}_j, \text{output}_j)$} for $j=1,\ldots,J$, where $J$ ranges between 5 and 10 and where both the input and output are strings.
\item
We form the following prompt: \texttt{$\text{input}_1 \diamond \text{output}_1 \oplus \text{input}_2 \diamond \text{output}_2 \oplus \ldots \oplus \text{input}_J \diamond \text{output}_J \oplus \text{new\_input}$}, where $\diamond$ denotes a newline character and $\oplus$ denotes a double newline character.
\item
To perform inference on a new input, we condition the LLM on the prompt and sample continuations of the prompt up until the next double newline character.
\end{enumerate}
All of our prompts are included in Figures~\ref{fig:prompt_cqgen}, \ref{fig:prompt_agreement}, and~\ref{fig:prompt_revision}. The contextual version used for QReCC are in Figures~\ref{fig:prompt_cqgen_contextual}, \ref{fig:prompt_agreement_contextual}, and~\ref{fig:prompt_revision_contextual}.

\paragraph{Model statistics}

We implemented most parts of RARR with the PALM model which has 540B parameters. We prompted PALM without any training or finetuning. We used a TPU  v2-128 to run inference with PALM.

We manually wrote our prompts by eye-balling quality on a dozen of examples from a separate validation set.
We tune our hyperparameters on the validation set as well. We used sampling temperature 0.7 for all generation tasks. For each input text, we sample 3 question generations, and for each question we retrieve 5 results. For agreement gate and editing, we only sample 1 generation. We reject an editing if the edit distance is more than 50 characters or more than half of the original text length.

\section{Details on the dataset}

As explained in Section~\ref{sec:eval-setup}, we generated 150 development and 150 test passages for each of the 6 combinations of dataset and model: (NQ, PaLM), (SQA, PaLM), (QReCC, LaMDA), (NQ, GPT-3), (SQA, GPT-3), (QReCC, GPT-3).
Figures~\ref{fig:prompt_passage_palm}, \ref{fig:prompt_passage_gpt3}, \ref{fig:prompt_passage_lamda},  and~\ref{fig:prompt_passage_gpt3_conv} are the few-shot prompts used to generate the passages.

Following the corresponding datasets, all generated passages are in English. The authors have manually looked through most of the data and found no personal identifiers.

\input{fig_prompts}

%% file: tab_ablation_results_upper.tex
\begin{table*}[!t]
\centering\small\setlength{\tabcolsep}{5pt}
\adjustbox{max width=\textwidth}{%
\begin{tabular}{
@{}l
r@{ $\to$ }rcc
r@{ $\to$ }rcc
r@{ $\to$ }rcc
@{}}
\toprule
 & \multicolumn{4}{c}{PaLM outputs on NQ}
 & \multicolumn{4}{c}{PaLM outputs on SQA}
 & \multicolumn{4}{c}{LaMDA outputs on QReCC}
 \\
 \cmidrule(lr){2-5} \cmidrule(lr){6-9} \cmidrule(lr){10-13}
 Model
 & \multicolumn{2}{c}{$\attrauto{}$} & $\preslev{}$ & \multicolumn{1}{c}{$\HM{}$}
 & \multicolumn{2}{c}{$\attrauto{}$} & $\preslev{}$ & \multicolumn{1}{c}{$\HM{}$}
 & \multicolumn{2}{c}{$\attrauto{}$} & $\preslev{}$ & \multicolumn{1}{c@{}}{$\HM{}$}
 \\
\midrule
Full \approach{}
& 45.6 & 54.9 & 89.6 & 68.1
& 37.6 & 45.1 & 89.9 & 60.0
& 18.8 & 29.4 & 80.2 & 43.1
\\ %
qgen 62B, editor 540B
& 45.9 & 54.6 & 87.8 & 67.4
& 37.0 & 40.5 & 90.0 & 55.9
& 15.8 & 28.4 & 76.1 & 41.4
\\
qgen 62B, editor 62B
& 45.9 & 49.9 & 91.0 & 64.4
& 37.0 & 38.3 & 93.0 & 54.2
& 15.8 & 21.9 & 71.6 & 33.5
\\ %
GPT-3
& 44.3 & 55.0 & 90.6 & \textbf{68.5}
& 38.6 & 46.6 & 89.3 & \textbf{61.2}
& 18.3 & 28.6 & 89.8 & \textbf{43.4}
\\
\bottomrule
\end{tabular}
}
\caption{\textbf{Additional ablation results.} We report the automatic metrics: $\attrauto$, $\preslev$, and harmonic mean between the two ($\HM{}$). We show auto-AIS scores both before and after editing (before $\rightarrow$ edit), with respect to the attribution report $A$ produced by the model.
}
\label{tab:ablation_results_upper}
\end{table*}

%% file: tab_gpt3_passages_results.tex
\begin{table*}[!t]
\centering\small
\adjustbox{max width=\textwidth}{%
\begin{tabular}{
@{}l
r@{ $\to$ }rcc
r@{ $\to$ }rcc
r@{ $\to$ }rcc
@{}}
\toprule
 & \multicolumn{4}{c}{GPT-3 outputs on NQ}
 & \multicolumn{4}{c}{GPT-3 outputs on SQA}
 & \multicolumn{4}{c}{GPT-3 outputs on QReCC}
 \\
 \cmidrule(lr){2-5} \cmidrule(lr){6-9} \cmidrule(lr){10-13}
 Model
 & \multicolumn{2}{c}{$\attrauto{}$} & $\preslev{}$ & \multicolumn{1}{c}{$\HM{}$}
 & \multicolumn{2}{c}{$\attrauto{}$} & $\preslev{}$ & \multicolumn{1}{c}{$\HM{}$}
 & \multicolumn{2}{c}{$\attrauto{}$} & $\preslev{}$ & \multicolumn{1}{c@{}}{$\HM{}$}
 \\
\midrule
EFEC
& 48.3 & 66.8 & 41.5 & 51.2
& 32.6 & 50.6 & 29.4 & 37.2
& 26.4 & 53.1 & 39.0 & 44.9
\\
LaMDA
& 36.2 & 61.1 & 45.9 & 52.4
& 22.3 & 27.3 & 43.3 & 33.5
& 19.0 & 33.9 & 28.3 & 30.8
\\
PaLM \approach{}
& 48.3 & 57.2 & 89.6 & 69.8
& 32.6 & 36.3 & 91.6 & 52.0
& 26.4 & 31.1 & 87.7 & \textbf{45.9}
\\
GPT-3 \approach{}
& 48.0 & 59.3 & 91.8 & \textbf{72.0}
& 34.7 & 37.0 & 91.8 & \textbf{52.8}
& 23.2 & 25.3 & 89.7 & 39.5
\\
\bottomrule
\end{tabular}
}
\caption{\textbf{Results on passages from GPT-3.} We report the automatic metrics: $\attrauto$, $\preslev$, and harmonic mean between the two ($\HM{}$). We show auto-AIS scores both before and after editing (before $\rightarrow$ edit), with respect to the attribution report $A$ produced by the model.
The results show similar trends as the results on passages from PaLM and LaMDA in Table~\ref{tab:main_results}.
}
\label{tab:gpt3_passages_results}
\end{table*}

%% file: tab_eli5_and_summeval.tex
\begin{table}[!t]
\centering\small
\adjustbox{max width=\columnwidth}{%
\begin{tabular}{
@{}l
r@{ $\to$ }rcc
@{}}
\toprule
 Model
 & \multicolumn{2}{c}{$\attrauto{}$} & $\preslev{}$ & $\HM{}$
 \\ \midrule
\multicolumn{5}{c}{\textbf{SummEval}}
\\
EFEC 
& 17.9 & 34.6 & 20.9 & 26.0 
\\
LaMDA
& 10.3 & 28.8 & 28.1 & 28.4 
\\
\approach{}
& 18.3 & 16.9 & 92.9 & 28.6 
\\
\midrule
\multicolumn{5}{c}{\textbf{ELI5}}
\\
EFEC 
& 18.2 & 41.2 & 17.2 & 24.2 
\\
LaMDA
& 19.9 & 40.1 & 31.2 & 35.1 
\\
\approach{}
& 18.5 & 18.9 & 97.2 & 31.7
\\

\bottomrule
\end{tabular}
}
\caption{\textbf{Results on ELI5 and SummEval.}}
\label{tab:eli5_and_summeval}
\end{table}

%% file: tab_mmlu.tex
\begin{table}[!t]
\centering\small
\adjustbox{max width=\columnwidth}{%
\begin{tabular}{
@{}l
r@{ $\to$ }rcc
@{}}
\toprule
 & \multicolumn{4}{c}{RARR}
 \\
MMLU Category
 & \multicolumn{2}{c}{$\attrauto{}$} & $\preslev{}$ & $\HM{}$
 \\ \midrule
\text{Humanities} 
& 26.6 & 29.6 & 6.6 & 45.0 
\\
\text{Social Sciences}
& 35.5 & 40.7 & 7.6 & 56.5
\\
\text{STEM}
& 37.8 & 41.5 & 7.2 & 57.4 
\\
\text{Other}
& 36.9 & 41.7 & 7.1 & 57.6 
\\ 
\bottomrule
\end{tabular}
}
\caption{\textbf{\approach{} results on MMLU.}}
\label{tab:mmlu}
\end{table}

%% file: fig_human_vs_auto_eval.tex
\begin{figure}[!t]
\centering
\includegraphics[width=\columnwidth]{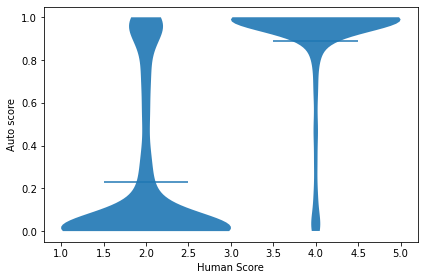}
\caption{
Violin plot illustrating the strong correlation between human AIS and auto-AIS labels on our NQ benchmark. Pearson correlation is 0.74 (N=450). $y$-axis is auto-AIS score, the two violins correspond to a human label of 0 or 1.
}\label{fig:human-vs-auto-eval}
\end{figure}

%% file: 7a_human_eval.tex
\section{Details on human evaluation}
\label{sec:app-human-eval}

To end-goal of \approach{} is to improve the \textit{attribution} of generation models through post-editing while \textit{preserving} the original intent.
Attribution and preservation are both subjective properties that may change with even small edits.
In the main paper, we present two automatic metrics to conveniently gauge these properties, but rely on a human evaluation as the gold standard.
In this section, we describe how we conducted the human evaluation and what instructions and examples annotators were provided.

\paragraph{Rater recruitment and training}
We engaged with a vendor supplier of full-time crowd workers to recruit human annotators for our task.
Annotators were asked to review the instructions below and were provided direct feedback on their responses during the pilot annotation runs.
We had 3 annotators rate each example in the pilot phase to measure inter-annotator agreement, and had a single rater annotate each example afterwards.

\subsection{Instructions: Overview}

In this task you will evaluate the quality of text generated by a system (\textbf{the ``passage''}) based on how well it represents information from multiple pieces of \textbf{``evidence''}.

We will be using two categories to evaluate the quality of the passage: \textbf{Attribution} and \textbf{Intent Similarity}. You will evaluate these categories in succession. In some tasks, you will only evaluate Attribution. The task interface will guide you through the flow; you can also see the overall task flow in the diagram below. 

\textbf{Note:} The passage may appear very fluent and well-formed, but still contain slight inaccuracies that are not easy to discern at first glance. Pay close attention to the text. Read it carefully as you would when proofreading.

\subsection{Instructions: Attribution}

\newcommand{\attrspan}[1]{\sethlcolor{lime}\hl{#1}}
\newcommand{\nattrspan}[1]{\sethlcolor{pink}\hl{#1}}

\begin{figure*}[!thp]
    \centering
    \includegraphics[width=\textwidth]{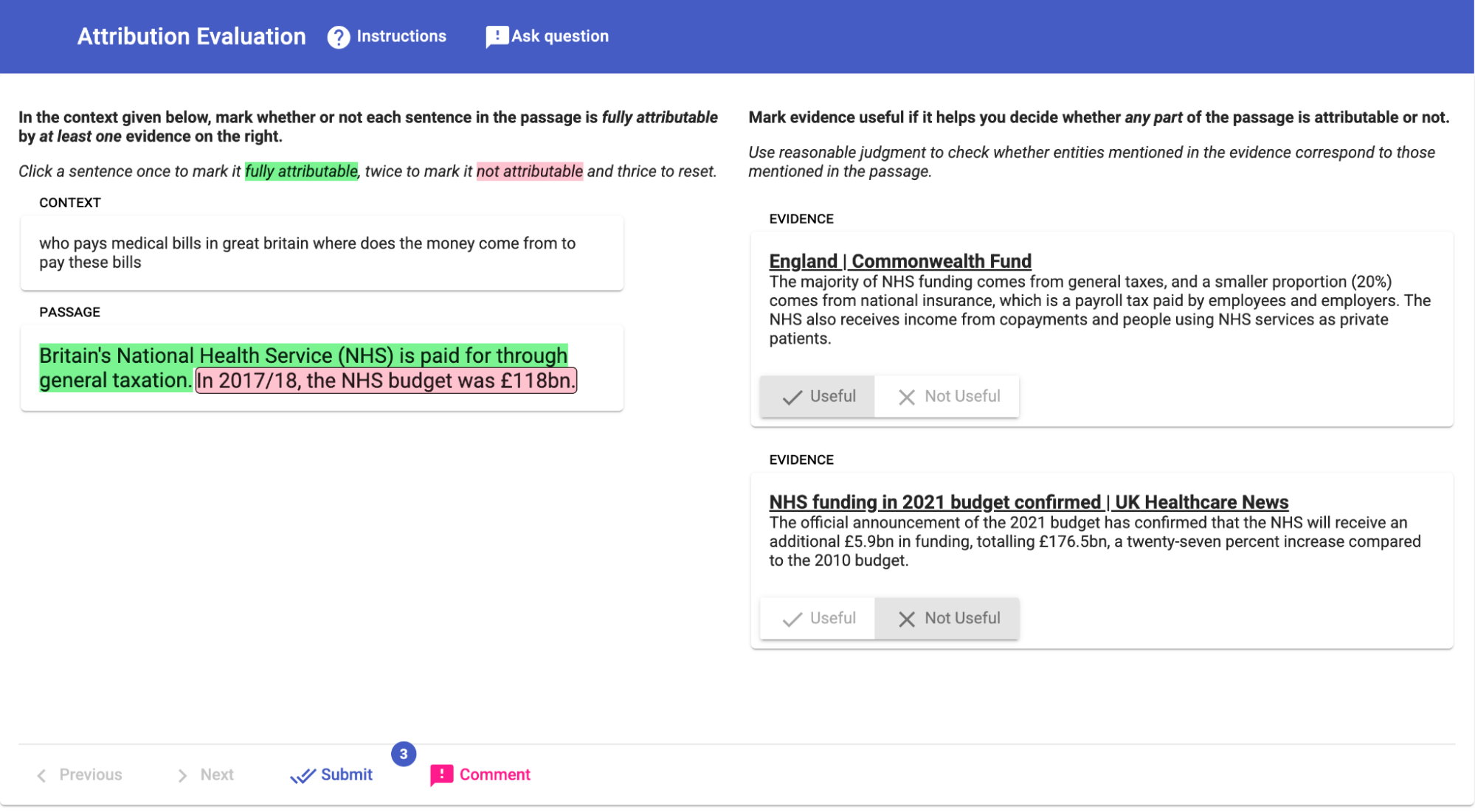}
    \caption{
    Screenshot of interface to annotate attribution at the sentence level.
    annotators were asked to mark sentences as being \attrspan{fully attributable} or \nattrspan{not fully attributable} by clicking each sentence, and rating each piece of evidence as being useful or not in helping determine attribution of the passage. Annotators were also presented with the context of the generation.
    }
    \label{fig:human-eval-attribution}
\end{figure*}
\begin{figure*}[!thp]
    \centering
    \includegraphics[width=\textwidth]{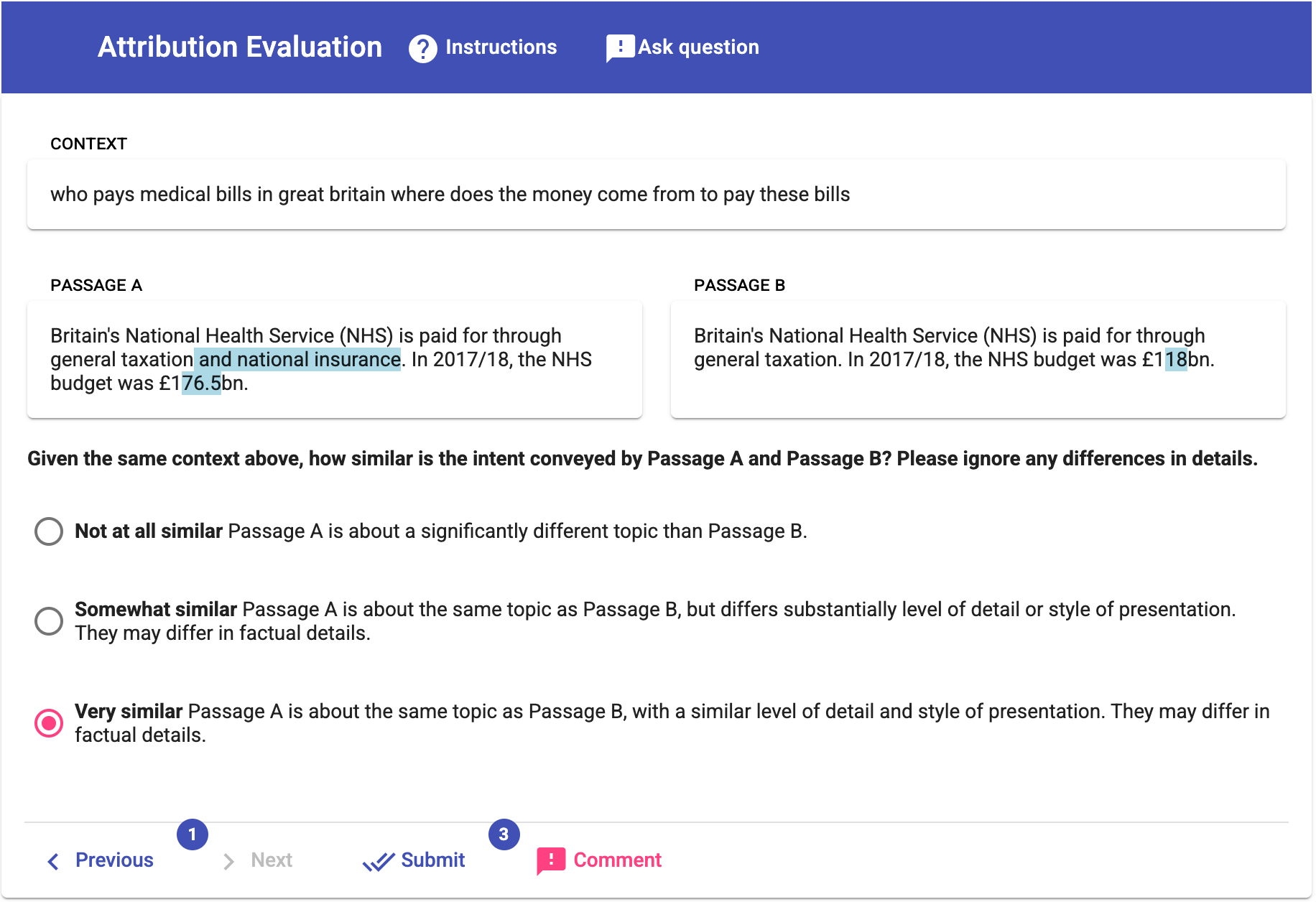}
    \caption{Screenshot of the preservation interface. Annotators are asked to read compare two passages and rate how similar the intent conveyed by the two passages is.}
    \label{fig:human-eval-preservation}
\end{figure*}

In this step, you will evaluate how much of the passage is attributable to one or more pieces of evidence (Figure \ref{fig:human-eval-attribution}).

In the interface, the passage of text and the context in which it was generated is shown on the left, and each piece of evidence is shown on the right. You will use all three (context, passage, evidence) to answer the following question for each sentence in the passage:
\textit{Is all of the information provided by this sentence fully supported by at least one piece of evidence?}

\paragraph{Determining the information provided by the sentence.}
Three points are key when determining information provided by the sentence:
\begin{enumerate}
\item The context and the other sentences of the passage are often critical in understanding the information provided by the sentence.
\item The context should only be used to understand the information provided by the sentence.
\item The evidence should be completely ignored for this step.
\end{enumerate}

Consider the following example:

\begin{quote}
\textit{Context:} who plays doug williams in days of our lives \\
\textit{Passage:} In the American daytime drama Days of Our Lives, Doug Williams and Julie Williams are portrayed by Bill Hayes and Susan Seaforth Hayes. \\
\end{quote}

In the above example, the meaning of the passage is clear even without seeing the query. But consider another example:

\begin{quote}
\textit{Context:} who plays doug williams in days of our lives \\
\textit{Passage:} he is played by Bill Hayes \\
\textit{Passage (interpreted):} \ul{Doug Williams} is played by Bill Hayes \ul{in days of our lives}
\end{quote}

In this case the pronoun ``he'' depends on the context, but it is clear that the  intended meaning of the passage can be reasonably interpreted as ``Doug Williams is played by Bill Hayes in days of our lives''. This interpretation is the ``information provided by the passage''. 

Pronouns such as he/she/it/they etc. are one case where context is needed to figure out the intended meaning of the system response. Here's another example (given with paraphrases of the information highlighted below):

\begin{quote}
\textit{Context:} when is the last time the us lost basketball at the olympics \\
\textit{Passage:} \ul{The last time they lost was in 2004}, when Argentina defeated the US 89–79. Most recently, they won gold in 2016. \\
\textit{Passage (interpreted):} The last time \ul{the United States lost basketball} at the Olympics was in 2004.
\end{quote}

The context should only be used to determine the information provided by the passage; at times, the passage may be about a slightly different topic than the context, for example:

\begin{quote}
\textit{Context:} the south west wind blows across nigeria between \\
\textit{Passage:} \ul{The Harmattan is a dry and dusty northeasterly trade wind that blows across West Africa} from December to March. It is very dusty because it blows across the Sahara.
\end{quote}

Here, the passage talks about a \textit{northeasterly wind}, while the context asks about a \textit{south-west wind}, but the passage can be fully understood.

In general, use your best judgment to determine the information provided by the passage. If the passage is hard to understand and you are unsure what the intended meaning of the passage is, \textit{mark the sentences as not attributed} and enter a comment with an explanation. As one example, take the following:

\begin{quote}
\textit{Context:} how many NBA championships did Michael Jordan win? \\
\textit{Passage:} it is the best team in the NBA
\end{quote}

\paragraph{Determining if the information accurately represents the evidence.}
Two points are key when determining whether the information accurately represents the evidence:
When interpreting a piece of evidence, use only the title and text of that specific evidence. Completely ignore the context, passage and all other evidence.
Check all the information in a sentence. If only some information is supported by the evidence, mark the sentence as not fully attributable.

Consider the following example:
\begin{quote}
\textit{Context:}
when did reba mcentire record back to god \\
\textit{Passage:}
\attrspan{Back to God was released by McEntire in 2017.} \\
\textit{Evidence:}
\ul{``Back to God'' is a song performed by American singer, Reba McEntire.} \ul{It was released} as the second single from her 2017 album, Sing it Now: Songs of Faith \& Hope, \ul{on January 20, 2017}.
\end{quote}
In the above example, it is reasonable to conclude that the evidence supports all the information in the passage, and we can mark the passage as being fully attributable. But consider another example:

\begin{quote}
\textit{Context:}
who won the womens 2017 ncaa basketball tournament \\
\textit{Passage:}
\nattrspan{South Carolina Gamecocks won the 2017 NCAA Women's Division I Basketball Tournament.} \\
\textit{Evidence:}
The \ul{South Carolina Gamecocks} defeated the Mississippi State Bulldogs, 67–55, to claim their first-ever \ul{national championship}.
\end{quote}

In this case, while the evidence also mentions the ``South Carolina Gamecocks'', it isn't clear that the national championship being mentioned is indeed the 2017 NCAA Women's Division I Basketball Tournament. The passage should be marked as not attributable.

Finally, when the passage contains multiple sentences, evaluate whether each sentence can be fully attributed to one or more pieces of evidence—it is possible for one sentence to be attributed while another is not. For example:

\begin{quote}
\textit{Context:}
who won the womens 2017 ncaa basketball tournament \\
\textit{Passage:}
\nattrspan{South Carolina Gamecocks won the 2017 NCAA Women's Division I Basketball Tournament.} \nattrspan{The final score is 67-55.} \attrspan{The championship game was held in Dallas, Texas.} \\
\textit{Evidence 1:}
The South Carolina Gamecocks defeated the Mississippi State Bulldogs, 67–55, to claim their first-ever national championship. \\
\textit{Evidence 2:}
The \ul{2017 NCAA Women's Division I Basketball Tournament} was played from Friday, March 17 to Sunday, April 2, 2017, with the Final Four played at the American Airlines Center in \ul{Dallas, Texas} on March 31 and April 2.
\end{quote}

The first two sentences cannot be attributed to either evidence for the same reason as the previous example, but the last sentence is fully supported by Evidence 2 and should be marked as attributed.

In general, you should use your best judgment in determining whether all of the information provided by the passage is ``an accurate representation of information in at least one evidence''. See Table \ref{tab:human-eval-attribution-examples} for additional examples.

We give the following final notes of guidance:
\begin{itemize}
\item  \textbf{Marking evidence as useful.} When reviewing each piece of evidence, mark it as useful if it helps you judge the attributability of any sentence, and mark it not useful if not. In the above example Evidence 1 is not useful because it didn't contain enough context to actually help you assess if the passage was attributable, but Evidence 2 was useful.
\item
\textbf{Contradicting evidence.} Mark a sentence as being attributed if any piece of evidence supports it: if two pieces of evidence contradict each other, but one of them supports the passage, mark the sentence as fully attributable.
\item
\textbf{More on the concept of ``accurate representation''.} We take as inspiration the journalist's conception of ``accurate representation''. For example, take \href{https://www.npr.org/about-npr/688139552/accuracy}{this excerpt on Accuracy in the NPR Ethics Handbook}: ``When quoting or paraphrasing anyone … consider whether the source would agree with the interpretation...'' In other words, if you had written the source document, consider whether you would view the system response as an accurate representation of information in that source document.
\end{itemize}

\begin{table*}
    \centering \small
    \begin{tabular}{@{}p{.25\textwidth} p{.45\textwidth} p{.25\textwidth}@{}}
    \toprule
        Context + Passage &
        Evidences &
        Notes \\ \midrule
        
\textit{Context: who played morticia in the addams family tv show}

\attrspan{The Addams Family is an American animated sitcom TV series.}
\nattrspan{It was first aired on NBC on September 24, 1973. Carolyn Jones played the role of Morticia.}
&
1/ The Addams Family (1973 TV series): The Addams Family is an American animated sitcom adaptation of the Charles Addams comic. The series was produced in 1973 and was rebroadcast the following season.

2/ The Addams Family (TV Series 1964–1966): When The Addams Family went off the air in 1966, network executives in charge of children's programming for NBC brought them back in 1973 for their own Saturday Morning cartoon show featuring the voices of Carolyn Jones from the original series.
&
While the evidence supports the show being aired in 1973, it doesn't specify the exact date (September 24, 1973).

Similarly, while the evidence mentions Carolyn Jones as being a voice actor, it doesn't say she played the role of Mortica.
\\      \midrule
\textit{
Context: when will the la sagrada familia be finished
}

\attrspan{The La Sagrada Familia is a large Roman Catholic church in Barcelona. It is designed by Antoni Gaudi.}
\nattrspan{It started construction in 1882, and the construction is still going on.}
\nattrspan{The estimated date to finish is 2026.}
&

1/ Sagrada Família - Wikipedia: The Basílica i Temple Expiatori de la Sagrada Família is a church in the Eixample district of Barcelona, Catalonia, Spain, and is currently the
largest unfinished Roman Catholic church.

2/ Find Out Sagrada Familia's Expected Finish Date: Visiting the breathtaking Sagrada Familia today also means witnessing the slow progress towards the completion of the project. Sagrada Familia is now  expected to be completed in 2026, the centenary of Gaudi's death.
It's a reasonable inference that La Sagrada Familia is the same as Sagrada Familia, even though the names differ slightly.
&
While Evidence 2 mentions Gaudi, it isn't clear this is a reference to Antoni Gaudi and further doesn't say that he designed the church.
\\
    \bottomrule
    \end{tabular}
    \caption{Additional examples for annotating attribution.}
    \label{tab:human-eval-attribution-examples}
\end{table*}

\subsection{Instructions: Intent Similarity}
In this step, you will evaluate how much similar the passage is to another passage (Figure \ref{fig:human-eval-preservation}).

In the interface, the passage A and passage B are both text generated by a system—given the same context in which it was generated. You will use all three (context, passage A, passage B) to answer the following question:
\textit{
How similar is the intent expressed by Passage A and Passage B? Please ignore any differences in details.
}

Two points are key when determining whether the two passages convey the same intent:
\begin{enumerate}
    \item  Judge the similarity solely based on the similarity in the type and quantity of information provided by each passage.
    \item Ignore any differences in factual details between the two passages.
\end{enumerate}

Consider the following examples:

\begin{quote}
\textit{Context:}
who pays medical bills in great britain where does the money come from to pay these bills \\
\textit{Passage A:}
Britain's National Health Service (NHS) is paid for through general taxation \ul{and national insurance}. In 2017/18, the NHS budget was £\ul{176.5bn}. \\
\textit{Passage B:}
Britain's National Health Service (NHS) is paid for through general taxation. In 2017/18, the NHS budget was £\ul{118bn}. \\
\textit{Rating:} \ul{Very similar.} Passage A is about the same topic as Passage B, with a similar level of detail and style of presentation. They may differ in factual details.
\end{quote}

The above example should be rated ``very similar'' because both passages include information about (1) how the NHS is paid for, and (2) what its budget in 2017/18 was, though they differ in their actual answers to these questions.

\begin{quote}
\textit{Context:}
who is the owner of reading football club \\
\textit{Passage A:}
Reading's owner is \ul{Yongge Dai}. Yongge \ul{Dai is also the president of Chinese company Dai Yongge Real Estate. Yongge's son, Dai Xiu Li, is Reading's vice-president.} \\
\textit{Passage B:}
Reading's owner is \ul{Dai Yongge}. Yongge's \ul{brother and sister pair behind the Reading FC takeover---Dai Yongge and Dai Xiu Li---has made their fortune through a massive property empire. Mr Dai, has been the chairman of Renhe Commercial since 1999, which is an organisation owned by his sister behind a vast network of underground shopping centres in China}. \\
\textit{Rating:}
\ul{Somewhat similar.} Passage A is about the same topic as Passage B, but differs substantially in level of detail or style of presentation. They may differ in factual details.
\end{quote}

The above example should be rated ``somewhat similar'' because both passages are still about the same topic—Reading's owner— but differ substantially in the information they discuss: Passage A includes information about (1a) who Reading's owner is, (2a) which company they are the president of and (3a) who their vice-president is. In contrast, while Passage B shares information about (1a), it also includes information about (2b) how the Reading owner made their fortune, (3b) their company position and how long they held it for and (4b) what the company also owns.

\begin{quote}
\textit{Context:}
what is the numbers of total elected member of indian parliment in present time \\
\textit{Passage A:}
The total number of elected members of the \ul{Lok} Sabha is \ul{543}. \\
\textit{Passage B:}
The total number of elected members of the \ul{Rajya} Sabha is \ul{238}. \\
\textit{Rating:}
\ul{Not at all similar.} Passage A is about a significantly different topic than Passage B.
\end{quote}

Even though the passages look very similar, the above example should be rated ``not at all similar'' because the two passages are about significantly different topics: ``the Lok Sabha'' vs ``the Rajya Sabha''.

%% file: fig_prompts.tex
{ %

\newcommand{\pWidth}{5.8in}
\newcounter{pCount}
\newcommand{\pStart}{\setcounter{pCount}{1}}
\newcommand{\pRow}[1]{\texttt{\tiny\color{gray}\arabic{pCount}}&\texttt{#1}\stepcounter{pCount}\\}
\newcommand{\pVar}[1]{\textbf{\{#1\}}}
\newcommand{\pBlank}{\_\_\_\_\_}

\begin{figure*}[p]
\begin{center}\scriptsize
\begin{tabular}{|rp{\pWidth}|}\hline\pStart&\\
\pRow{[web] I will check things you said and ask questions.}
\pRow{}
\pRow{(1) You said: Your nose switches back and forth between nostrils. When you sleep, you switch about every 45 minutes.  This is to prevent a buildup of mucus. It's called the nasal cycle.}
\pRow{To verify it,}
\pRow{a) I googled: Does your nose switch between nostrils?}
\pRow{b) I googled: How often does your nostrils switch?}
\pRow{c) I googled: Why does your nostril switch?}
\pRow{d) I googled: What is nasal cycle?}
\pRow{}
\pRow{(2) You said: The Stanford Prison Experiment was conducted in the basement of Encina Hall, Stanford's psychology building.}
\pRow{To verify it,}
\pRow{a) I googled: Where was Stanford Prison Experiment was conducted?}
\pRow{}
\pRow{(3) You said: The Havel-Hakimi algorithm is an algorithm for converting the adjacency matrix of a graph into its adjacency list. It is named after Vaclav Havel and Samih Hakimi.}
\pRow{To verify it,}
\pRow{a) I googled: What does Havel-Hakimi algorithm do?}
\pRow{b) I googled: Who are Havel-Hakimi algorithm named after?}
\pRow{}
\pRow{(4) You said: "Time of My Life" is a song by American singer-songwriter Bill Medley from the soundtrack of the 1987 film Dirty Dancing. The song was produced by Michael Lloyd.}
\pRow{To verify it,}
\pRow{a) I googled: Who sings "Time of My Life"?}
\pRow{b) I googled: Which film is "Time of My Life" from?}
\pRow{c) I googled: Who produced the song "Time of My Life"?}
\pRow{}
\pRow{(5) You said: Kelvin Hopins was suspended from the Labor Party due to his membership in the Conservative Party.}
\pRow{To verify it,}
\pRow{a) I googled: Why was Kelvin Hopins suspended from Labor Party?}
\pRow{}
\pRow{(6) You said: Social work is a profession that is based in the philosophical tradition of humanism. It is an intellectual discipline that has its roots in the 1800s.}
\pRow{To verify it,}
\pRow{a) I googled: What philosophical tradition is social work based on?}
\pRow{b) I googled: What year does social work has its root in?}
\pRow{}
\pRow{(7) You said: \pVar{text}}
\pRow{To verify it,}
\pRow{\pBlank}
&\\\hline\end{tabular}
\caption{Few-shot prompt for query generation. To increase diversity and coverage, we sample the model three times and combine the resulting lists of queries.}
\label{fig:prompt_cqgen}
\end{center}
\end{figure*}

\begin{figure*}[p]
\begin{center}\scriptsize
\begin{tabular}{|rp{\pWidth}|}\hline\pStart&\\
\pRow{[web] I will check some things you said.}
\pRow{}
\pRow{(1) You said: Your nose switches back and forth between nostrils. When you sleep, you switch about every 45 minutes.  This is to prevent a buildup of mucus. It's called the nasal cycle.}
\pRow{I checked: How often do your nostrils switch?}
\pRow{I found this article: Although we don't usually notice it, during the nasal cycle one nostril becomes congested and thus contributes less to airflow, while the other becomes decongested. On average, the congestion pattern switches about every 2 hours, according to a small 2016 study published in the journal PLOS One.}
\pRow{Your nose's switching time is about every 2 hours, not 45 minutes.}
\pRow{This disagrees with what you said. }
\pRow{}
\pRow{(2) You said: The Little House books were written by Laura Ingalls Wilder. The books were published by HarperCollins.}
\pRow{I checked: Who published the Little House books?}
\pRow{I found this article: These are the books that started it all -- the stories that captured the hearts and imaginations of children and young adults worldwide. Written by Laura Ingalls Wilder and published by HarperCollins, these beloved books remain a favorite to this day.}
\pRow{The Little House books were published by HarperCollins.}
\pRow{This agrees with what you said.}
\pRow{}
\pRow{(3) You said: The Stanford Prison Experiment was conducted in the basement of Jordan Hall, Stanford's psychology building.}
\pRow{I checked: Where was Stanford Prison Experiment conducted?}
\pRow{I found this article: Carried out August 15-21, 1971 in the basement of Jordan Hall, the Stanford Prison Experiment set out to examine the psychological effects of authority and powerlessness in a prison environment.}
\pRow{The Stanford Prison Experiment was conducted in Jordan Hall.}
\pRow{This agrees with what you said.}
\pRow{}
\pRow{(4) You said: Social work is a profession that is based in the philosophical tradition of humanism. It is an intellectual discipline that has its roots in the 1800s.}
\pRow{I checked: When did social work have its roots?}
\pRow{I found this article: The Emergence and Growth of the Social work Profession<br><br> Social work's roots were planted in the 1880s, when charity organization societies (COS) were created to organize municipal voluntary relief associations and settlement houses were established.}
\pRow{Social work has its roots in the 1880s, not 1800s.}
\pRow{This disagrees with what you said.}
\pRow{}
\pRow{(5) You said: The Havel-Hakimi algorithm is an algorithm for converting the adjacency matrix of a graph into its adjacency list. It is named after Vaclav Havel and Samih Hakimi.}
\pRow{I checked: What is the Havel-Hakimi algorithm?}
\pRow{I found this article: The Havel-Hakimi algorithm constructs a special solution if a simple graph for the given degree sequence exists, or proves that one cannot find a positive answer. This construction is based on a recursive algorithm. The algorithm was published by Havel (1955), and later by Hakimi (1962).}
\pRow{Havel-Hakimi algorithm is for constructing a special solution if a simple graph for the given degree sequence exists, or proving that one cannot find a positive answer, not converting the adjacency matrix of a graph into its adjacency list.}
\pRow{This disagrees with what you said.}
\pRow{}
\pRow{(6) You said: "Time of My Life" is a song by American singer-songwriter Bill Medley from the soundtrack of the 1987 film Dirty Dancing. The song was produced by Michael Lloyd.}
\pRow{I checked: Who was the producer of "(I've Had) The Time of My Life"?}
\pRow{I found this article: On September 8, 2010, the original demo of this song, along with a remix by producer Michael Lloyd, was released as digital files in an effort to raise money for the Patrick Swayze Pancreas Cancer Resarch Foundation at Stanford University.}
\pRow{"Time of My Life" was produced by Michael Lloyd.}
\pRow{This agrees with what you said.}
\pRow{}
\pRow{(7) You said: Kelvin Hopins was suspended from the Labor Party because he had allegedly sexually harassed and behaved inappropriately towards a Labour Party activist, Ava Etemadzadeh.}
\pRow{I checked: Why was Kelvin Hopins suspeneded from the Labor Party? }
\pRow{I found this article: A former Labour MP has left the party before an inquiry into sexual harassment allegations against him was able to be concluded, the party has confirmed. Kelvin Hopkins was accused in 2017 of inappropriate physical contact and was suspended by the Labour party pending an investigation.This agrees with what you said.}
\pRow{Kelvin Hopins was suspended because he had allegedly sexually harassed and behaved inappropriately towards a Labour Party activist, Ava Etemadzadeh.}
\pRow{This agrees with what you said.}
\pRow{}
\pRow{(8) You said: In the battles of Lexington and Concord, the British side was led by General Thomas Smith.}
\pRow{I checked: Who led the British side in the battle of Lexington and Concord?}
\pRow{I found this article: Interesting Facts about the Battles of Lexington and Concord. The British were led by Lieutenant Colonel Francis Smith. There were 700 British regulars.}
\pRow{The British side was led by Lieutenant Colonel Francis Smith, not General Thomas Hall.}
\pRow{This disagrees with what you said.}
\pRow{}
\pRow{(9) You said: \pVar{text}}
\pRow{I checked: \pVar{query}}
\pRow{I found this article: \pVar{evidence}}
\pRow{\pBlank}
&\\\hline\end{tabular}
\caption{Few-shot prompt for the agreement model, which uses chain-of-thought prompting.}
\label{fig:prompt_agreement}
\end{center}
\end{figure*}

\begin{figure*}[p]
\begin{center}\scriptsize
\begin{tabular}{|rp{\pWidth}|}\hline\pStart&\\
\pRow{[web] I will fix some things you said.}
\pRow{}
\pRow{(1) You said: Your nose switches back and forth between nostrils. When you sleep, you switch about every 45 minutes.  This is to prevent a buildup of mucus. It's called the nasal cycle.}
\pRow{I checked: How often do your nostrils switch?}
\pRow{I found this article: Although we don't usually notice it, during the nasal cycle one nostril becomes congested and thus contributes less to airflow, while the other becomes decongested. On average, the congestion pattern switches about every 2 hours, according to a small 2016 study published in the journal PLOS One.}
\pRow{This suggests 45 minutes switch time in your statement is wrong.}
\pRow{My fix: Your nose switches back and forth between nostrils. When you sleep, you switch about every 2 hours. This is to prevent a buildup of mucus. It's called the nasal cycle.}
\pRow{}
\pRow{(2) You said: In the battles of Lexington and Concord, the British side was led by General Thomas Hall.}
\pRow{I checked: who led the British side in the battle of Lexington and Concord?}
\pRow{I found this article: Interesting Facts about the Battles of Lexington and Concord. The British were led by Lieutenant Colonel Francis Smith. There were 700 British regulars.}
\pRow{This suggests General Thomas Hall in your statement is wrong.}
\pRow{My fix: In the battles of Lexington and Concord, the British side was led by Lieutenant Colonel Francis Smith.}
\pRow{}
\pRow{(3) You said: The Stanford Prison Experiment was conducted in the basement of Encina Hall, Stanford's psychology building.}
\pRow{I checked: where was Stanford Prison Experiment conducted.}
\pRow{I found this article: Carried out August 15-21, 1971 in the basement of Jordan Hall, the Stanford Prison Experiment set out to examine the psychological effects of authority and powerlessness in a prison environment.}
\pRow{This suggests Encina Hall in your statement is wrong.}
\pRow{My fix: The Stanford Prison Experiment was conducted in the basement of Jordan Hall, Stanford's psychology building.}
\pRow{}
\pRow{(4) You said: Phoenix Mills Ltd., a diversified business conglomerate, was established in 1854. It has a history of over 160 years.}
\pRow{I checked: When was Phoenix Mills Ltd. founded?}
\pRow{I found this article: Phoenix Mills Ltd was incorporated in the year 1905. The company began their operations as a textile manufacturing company on 17.3 acres of land at Lower Parel in Mumbai. In the year 1959 the company was listed in the Bombay Stock Exchange.}
\pRow{This suggests the year of establishment 1854 in your statement is wrong.}
\pRow{My fix: Phoenix Mills Ltd., a diversified business conglomerate, was established in 1905. It has a history of over 160 years.}
\pRow{}
\pRow{(5) You said: The Havel-Hakimi algorithm is an algorithm for converting the adjacency matrix of a graph into its adjacency list. It is named after Vaclav Havel and Samih Hakimi.}
\pRow{I checked: What is the Havel-Hakimi algorithm?}
\pRow{I found this article: The Havel-Hakimi algorithm constructs a special solution if a simple graph for the given degree sequence exists, or proves that one cannot find a positive answer. This construction is based on a recursive algorithm. The algorithm was published by Havel (1955), and later by Hakimi (1962).}
\pRow{This suggests the Havel-Hakimi algorithm's functionality in your statement is wrong.}
\pRow{My fix: The Havel-Hakimi algorithm constructs a special solution if a simple graph for the given degree sequence exists, or proves that one cannot find a positive answer. It is named after Vaclav Havel and Samih Hakimi}
\pRow{}
\pRow{(6) You said: "Time of My Life" is a song by American singer-songwriter Bill Medley from the soundtrack of the 1987 film Dirty Dancing. The song was produced by Phil Ramone.}
\pRow{I checked: Who was the producer of "(I've Had) The Time of My Life"?}
\pRow{I found this article: On September 8, 2010, the original demo of this song, along with a remix by producer Michael Lloyd, was released as digital files in an effort to raise money for the Patrick Swayze Pancreas Cancer Resarch Foundation at Stanford University.}
\pRow{This suggests "Time of My Life" producer name in your statement is wrong.}
\pRow{My fix: "Time of My Life" is a song by American singer-songwriter Bill Medley from the soundtrack of the 1987 film Dirty Dancing. The song was produced by Michael Lloyd.}
\pRow{}
\pRow{(7) You said: Phoenix Market City Pune is located on 21 acres of prime property in Pune. It is spread across four levels with approximately 1.4 million square feet of built-up space. The mall is owned and operated by Phoenix Mills Limited.}
\pRow{I checked: What is the area of Phoenix Market City in Pune?}
\pRow{I found this article: Phoenix Market City was opened in January 2013 and has the distinction of being the largest mall in the city of Pune, with the area of 3.4 million square feet. It is located in the Viman Nagar area of Pune.}
\pRow{This suggests the 1.4 million square feet of built-up space in your statment is wrong.}
\pRow{My fix: Phoenix Market City Pune is located on 21 acres of prime property in Pune. It is spread across four levels with approximately 3.4 million square feet of built-up space. The mall is owned and operated by Phoenix Mills Limited.}
\pRow{}
\pRow{(8) You said: \pVar{text}}
\pRow{I checked: \pVar{query}}
\pRow{I found this article: \pVar{evidence}}
\pRow{This suggests \pBlank}
&\\\hline\end{tabular}
\caption{Few-shot prompt for the revision model, which uses chain-of-thought prompting.}
\label{fig:prompt_revision}
\end{center}
\end{figure*}

\begin{figure*}[p]
\begin{center}\scriptsize
\begin{tabular}{|rp{\pWidth}|}\hline\pStart&\\
\pRow{[web] I will read the context and check only the last thing you said by asking questions.}
\pRow{}
\pRow{(1) Context: Your nose switches back and forth between nostrils. When you sleep, you switch about every 45 minutes.}
\pRow{You said: This is to prevent a buildup of mucus. It's called the nasal cycle.}
\pRow{To verify what you just said,}
\pRow{a) I googled: Why does your nostril switch during sleep?}
\pRow{b) I googled: What is nasal cycle?}
\pRow{c) I googled: What is the nostril switching during sleep called?}
\pRow{}
\pRow{(2) Context: The Stanford Prison Experiment was conducted in the basement of Encina Hall, Stanford's psychology building.}
\pRow{You said: It is a psychological study to observe the behaviors of conflict and violence that happen between inmates and prisoners in real prisons.}
\pRow{To verify what you just said,}
\pRow{a) I googled: What type of experiment was the Stanford Prison Experiment?}
\pRow{b) I googled: What was the objective of the Stanford Prison Experiment?}
\pRow{}
\pRow{(3) Context: The Havel-Hakimi algorithm is an algorithm for converting the adjacency matrix of a graph into its adjacency list.}
\pRow{You said: It is named after Vaclav Havel and Samih Hakimi.}
\pRow{To verify what you just said,}
\pRow{a) I googled: Who are Havel-Hakimi algorithm named after?}
\pRow{}
\pRow{(4) Context: "Time of My Life" is a song by American singer-songwriter Bill Medley from the soundtrack of the 1987 film Dirty Dancing.}
\pRow{You said: The song was produced by Michael Lloyd in the same year.}
\pRow{To verify what you just said,}
\pRow{a) I googled: Who produced the song "Time of My Life"?}
\pRow{b) I googled: When was the song "Time of My Life" by Bill Medley produced?}
\pRow{}
\pRow{(5) Context: The Late Show with Stephen Colbert is an American late-night talk show hosted by Stephen Colbert, which premiered on September 8, 2015.}
\pRow{You said: Produced by Spartina Productions and CBS Television Studios, it is the second iteration of CBS' Late Show franchise.}
\pRow{To verify what you just said,}
\pRow{a) I googled: Who produces "The Late Show with Stephen Colbert"?}
\pRow{b) I googled: What are the iterations of CBS' Late Show franchise?}
\pRow{}
\pRow{(6) Context: Super Mario Sunshine was released on GameCube in 2002. In the game, Mario uses a tool strapped to his back called FLUDD, which stands for The Flash Liquidizer Ultra Dousing Device.}
\pRow{You said: It can be used to spray water at objects or enemies. This allows Mario to change his movements, kill enemies, or clean up hazards on the floor.}
\pRow{To verify what you just said,}
\pRow{a) I googled: What is the main function of FLUDD in Super Mario Sunshine?}
\pRow{b) I googled: What can FLUDD in Super Mario Sunshine be used on?}
\pRow{c) I googled: In Super Mario Sunshine, can Mario change movement with FLUDD?}
\pRow{d) I googled: In Super Mario Sunshine, can Mario kill enemies with FLUDD?}
\pRow{e) I googled: In Super Mario Sunshine, can Mario clean up hazards on the floor with FLUDD?}
\pRow{}
\pRow{(7) Context: \pVar{context}}
\pRow{You said: \pVar{text}}
\pRow{To verify what you just said,}
\pRow{\pBlank}
&\\\hline\end{tabular}
\caption{Contextual version of the query generation prompt. The prompt works well for dialog contexts from QReCC even though the few-shot examples are not formatted as such.}
\label{fig:prompt_cqgen_contextual}
\end{center}
\end{figure*}

\begin{figure*}[p]
\begin{center}\scriptsize
\begin{tabular}{|rp{\pWidth}|}\hline\pStart&\\
\pRow{[web] I will check some things you said.}
\pRow{}
\pRow{(1) Context: Your nose switches back and forth between nostrils. It's called the nasal cycle. This is to prevent a buildup of mucus.}
\pRow{You said: When you sleep, you switch about every 45 minutes.}
\pRow{I checked: How often do your nostrils switch?}
\pRow{I found this article: Although we don't usually notice it, during the nasal cycle one nostril becomes congested and thus contributes less to airflow, while the other becomes decongested. On average, the congestion pattern switches about every 2 hours, according to a small 2016 study published in the journal PLOS One.}
\pRow{Your nose's switching time is about every 2 hours, not 45 minutes.}
\pRow{This disagrees with what you said.}
\pRow{}
\pRow{(2) Context: The Little House books is a series of American children's novels.}
\pRow{You said: The books were published by HarperCollins.}
\pRow{I checked: Who published the Little House books?}
\pRow{I found this article: These are the books that started it all -- the stories that captured the hearts and imaginations of children and young adults worldwide. Written by Laura Ingalls Wilder and published by HarperCollins, these beloved books remain a favorite to this day.}
\pRow{The Little House books were published by HarperCollins.}
\pRow{This agrees with what you said.}
\pRow{}
\pRow{(3) Context: The Stanford Prison Experiment is a psychological study to observe the behaviors of conflict and violence that happen between inmates and prisoners in real prisons.}
\pRow{You said: It was conducted in the basement of Jordan Hall, Stanford's psychology building.}
\pRow{I checked: Where was Stanford Prison Experiment conducted?}
\pRow{I found this article: Carried out August 15-21, 1971 in the basement of Jordan Hall, the Stanford Prison Experiment set out to examine the psychological effects of authority and powerlessness in a prison environment.}
\pRow{The Stanford Prison Experiment was conducted in Jordan Hall.}
\pRow{This agrees with what you said.}
\pRow{}
\pRow{(4) Context: Social work is a profession that is based in the philosophical tradition of humanism.}
\pRow{You said: It is an intellectual discipline that has its roots in the 1800s.}
\pRow{I checked: When did social work have its roots?}
\pRow{I found this article: The Emergence and Growth of the Social work Profession<br><br> Social work's roots were planted in the 1880s, when charity organization societies (COS) were created to organize municipal voluntary relief associations and settlement houses were established.}
\pRow{Social work has its roots in the 1880s, not 1800s.}
\pRow{This disagrees with what you said.}
\pRow{}
\pRow{(5) Context: The Havel-Hakimi algorithm is named after Vaclav Havel and Samih Hakimi.}
\pRow{You said: It is an algorithm for converting the adjacency matrix of a graph into its adjacency list.}
\pRow{I checked: What is the Havel-Hakimi algorithm?}
\pRow{I found this article: The Havel-Hakimi algorithm constructs a special solution if a simple graph for the given degree sequence exists, or proves that one cannot find a positive answer. This construction is based on a recursive algorithm. The algorithm was published by Havel (1955), and later by Hakimi (1962).}
\pRow{Havel-Hakimi algorithm is for constructing a special solution if a simple graph for the given degree sequence exists, or proving that one cannot find a positive answer, not converting the adjacency matrix of a graph into its adjacency list.}
\pRow{This disagrees with what you said.}
\pRow{}
\pRow{(6) Context: "Time of My Life" is a song by American singer-songwriter Bill Medley from the soundtrack of the 1987 film Dirty Dancing. }
\pRow{You said: The song was produced by Michael Lloyd in the same year.}
\pRow{I checked: Who was the producer of "(I've Had) The Time of My Life"?}
\pRow{I found this article: On September 8, 2010, the original demo of this song, along with a remix by producer Michael Lloyd, was released as digital files in an effort to raise money for the Patrick Swayze Pancreas Cancer Resarch Foundation at Stanford University.}
\pRow{The song "Time of My Life" was produced by Michael Lloyd.}
\pRow{This agrees with what you said.}
\pRow{}
\pRow{(7) Context: Super Mario Sunshine was released on GameCube in 2002. In the game, Mario uses a tool strapped to his back called FLUDD.}
\pRow{You said: FLUDD stands for Functional Language in a Unified Design Discipline. It can be used to spray water at objects or enemies. This allows Mario to change his movements, kill enemies, or clean up hazards on the floor.}
\pRow{I checked: What does FLUDD stands for in Super Mario Sunshine?}
\pRow{I found this article: The Flash Liquidizer Ultra Dousing Device, abbreviated and better known as FLUDD or F.L.U.D.D., is a multipurpose water pack from Super Mario Sunshine invented by Professor Elvin Gadd, indicated by the Gadd Science, Incorporated logo at the base of its nozzle exclusively during the cutscene at Pinna Park.}
\pRow{In Super Mario Sunshine, FLUDD stands for the Flash Liquidizer Ultra Dousing Device, not Functional Language in a Unified Design Discipline.}
\pRow{This disagrees with what you said.}
\pRow{}
\pRow{(8) Context: \pVar{context}}
\pRow{You said: \pVar{text}}
\pRow{I checked: \pVar{query}}
\pRow{I found this article: \pVar{evidence}}
\pRow{\pBlank}
&\\\hline\end{tabular}
\caption{Contextual version of the agreement model prompt.}
\label{fig:prompt_agreement_contextual}
\end{center}
\end{figure*}

\begin{figure*}[p]
\begin{center}\scriptsize
\begin{tabular}{|rp{\pWidth}|}\hline\pStart&\\
\pRow{[web] I will fix some things you said.}
\pRow{}
\pRow{(1) Context: Your nose switches back and forth between nostrils. It's called the nasal cycle. This is to prevent a buildup of mucus.}
\pRow{You said: When you sleep, you switch about every 45 minutes.}
\pRow{I checked: How often do your nostrils switch?}
\pRow{I found this article: Although we don't usually notice it, during the nasal cycle one nostril becomes congested and thus contributes less to airflow, while the other becomes decongested. On average, the congestion pattern switches about every 2 hours, according to a small 2016 study published in the journal PLOS One.}
\pRow{This suggests 45 minutes switch time in your statement is wrong.}
\pRow{My fix: When you sleep, you switch about every 2 hours.}
\pRow{}
\pRow{(2) Context: The Little House books is a series of American children's novels.}
\pRow{You said: The books were published by Amberjack Publishing.}
\pRow{I checked: Who published the Little House books?}
\pRow{I found this article: These are the books that started it all -- the stories that captured the hearts and imaginations of children and young adults worldwide. Written by Laura Ingalls Wilder and published by HarperCollins, these beloved books remain a favorite to this day.}
\pRow{This suggests Amberjack Publishing in your statement is wrong.}
\pRow{My fix: The books were published by HarperCollins.}
\pRow{}
\pRow{(3) Context: The Stanford Prison Experiment is a psychological study to observe the behaviors of conflict and violence that happen between inmates and prisoners in real prisons.}
\pRow{You said: It was conducted in the basement of Encina Hall, Stanford's psychology building.}
\pRow{I checked: where was Stanford Prison Experiment conducted.}
\pRow{I found this article: Carried out August 15-21, 1971 in the basement of Jordan Hall, the Stanford Prison Experiment set out to examine the psychological effects of authority and powerlessness in a prison environment.}
\pRow{This suggests Encina Hall in your statement is wrong.}
\pRow{My fix: It was conducted in the basement of Jordan Hall, Stanford's psychology building.}
\pRow{}
\pRow{(4) Context: The Havel-Hakimi algorithm is named after Vaclav Havel and Samih Hakimi.}
\pRow{You said: It is an algorithm for converting the adjacency matrix of a graph into its adjacency list.}
\pRow{I checked: What is the Havel-Hakimi algorithm?}
\pRow{I found this article: The Havel-Hakimi algorithm constructs a special solution if a simple graph for the given degree sequence exists, or proves that one cannot find a positive answer. This construction is based on a recursive algorithm. The algorithm was published by Havel (1955), and later by Hakimi (1962).}
\pRow{This suggests the Havel-Hakimi algorithm's functionality in your statement is wrong.}
\pRow{My fix: It constructs a special solution if a simple graph for the given degree sequence exists, or proves that one cannot find a positive answer.}
\pRow{}
\pRow{(5) Context: "Time of My Life" is a song by American singer-songwriter Bill Medley from the soundtrack of the 1987 film Dirty Dancing.}
\pRow{You said: The song was produced by Phil Ramone in the same year.}
\pRow{I checked: Who was the producer of "(I've Had) The Time of My Life"?}
\pRow{I found this article: On September 8, 2010, the original demo of this song, along with a remix by producer Michael Lloyd, was released as digital files in an effort to raise money for the Patrick Swayze Pancreas Cancer Resarch Foundation at Stanford University.}
\pRow{This suggests "Time of My Life" producer name in your statement is wrong.}
\pRow{My fix: The song was produced by Michael Lloyd in the same year.}
\pRow{}
\pRow{(6) Context: Phoenix Market City Pune is located on 21 acres of prime property in Pune. The mall is owned and operated by Phoenix Mills Limited.}
\pRow{You said: It is spread across four levels with approximately 1.4 million square feet of built-up space.}
\pRow{I checked: What is the area of Phoenix Market City in Pune?}
\pRow{I found this article: Phoenix Market City was opened in January 2013 and has the distinction of being the largest mall in the city of Pune, with the area of 3.4 million square feet. It is located in the Viman Nagar area of Pune.}
\pRow{This suggests the 1.4 million square feet of built-up space in your statment is wrong.}
\pRow{My fix: It is spread across four levels with approximately 3.4 million square feet of built-up space.}
\pRow{}
\pRow{(7) Context: \pVar{context}}
\pRow{You said: \pVar{text}}
\pRow{I checked: \pVar{query}}
\pRow{I found this article: \pVar{evidence}}
\pRow{This suggests \pBlank}
&\\\hline\end{tabular}
\caption{Contextual version of the revision model prompt.}
\label{fig:prompt_revision_contextual}
\end{center}
\end{figure*}

\begin{figure*}[p]
\begin{center}\scriptsize
\begin{tabular}{|rp{\pWidth}|}\hline\pStart&\\
\pRow{[web] I will think step by step and answer your question.}
\pRow{}
\pRow{Question: is growing seedless cucumber good for a gardener with entomophobia}
\pRow{Explanation: Entomophobia is a fear of insects. Plants need insects to pollinate them. Seedless fruits such as seedless cucumbers do not require pollination, so seedless fruits do not require insects. This makes good for people with entomophobia. }
\pRow{Answer: Yes}
\pRow{}
\pRow{Question: Who was british pm and viceroy during quit india movement?}
\pRow{Explanation: The Quit India Movement was launched in 8th August 1942. The british PM at that time was Winston Churchill. The british viceroy during the movement is Victor Hope, usually referred to as Lord Linlithgow. }
\pRow{Answer: Winston Churchil and Lord Linlithgow}
\pRow{}
\pRow{Question: Which year does game over man come out on netflix?}
\pRow{Explanation: Game Over, Man! is an action-comedy movie released in March 23, 2018, on Netflix. It's director is Kyle Newacheck. The movie stars Anders Holm, Adam DeVine, and Blake Anderson. }
\pRow{Answer: March 23, 2018}
\pRow{}
\pRow{Question: would it be very difficult for Nuno Gomes to dive to the Red Sea's deepest point?}
\pRow{Explanation: Nuno Gomes' deepest dive in the Red Sea to date is 317 metres. The Red Sea has a maximum depth of over 3,000 metres. So it would be difficult for Nuno Gomes to the deepest point of the Red sea.}
\pRow{Answer: Yes}
\pRow{}
\pRow{Question: Are chinchillas cold-blooded?}
\pRow{Explanation: Chinchillas are rodents, which are mammals. All mammals are warm-blooded.}
\pRow{Answer: No}
\pRow{}
\pRow{Question: what are the chemo drugs for gastric cancer?}
\pRow{Explanation: Chemotherapy can be given for gastric cancer before or after surgery. Usually you have a combination of 2 or 3 drugs. The most common types are: epirubicin, cisplatin and capecitabine (ECX) ; epirubicin, cisplatin and fluorouracil (ECF); or fluorouracil, folinic acid, oxaliplatin and docetaxel (FLOT).}
\pRow{Answer: epirubicin, cisplatin and capecitabine (ECX) ; epirubicin, cisplatin and fluorouracil (ECF); or fluorouracil, folinic acid, oxaliplatin and docetaxel (FLOT)}
\pRow{}
\pRow{Question: How many sacks does Clay Matthews have in his career?}
\pRow{Explanation: Clay Matthews III had 383 solo tackles, 136 assists, and 91.5 sacks. His father, Clay Matthews Jr., also palyed in NFL and had 69.5 sacks in his career. }
\pRow{Answer: 91.5 (Clay Matthews III) and 69.5 (Clay Matthews Jr.)}
\pRow{}
\pRow{Question: In the U.S., the most deaths are caused by}
\pRow{Explanation: The leading cause of death in the U.S. is heart disease, causing nearly 700,000 deaths annually. The second most common cause of deaths is cancer.}
\pRow{Answer: heart disease}
\pRow{}
\pRow{Question: the rank of indian economy in terms of nominal gdp is}
\pRow{Explanation: India's GDP is economy is \$2.94 trillion. It is the fifth-largest in the world. The top GDP contries are United States, China, Japan, Germany and India. }
\pRow{Answer: 5}
\pRow{}
\pRow{Question: \pVar{question}}
\pRow{Explanation: \pBlank}
&\\\hline\end{tabular}
\caption{The PaLM prompt for generating long-form answers to questions from NQ and SQA.}
\label{fig:prompt_passage_palm}
\end{center}
\end{figure*}

\begin{figure*}[p]
\begin{center}\scriptsize
\begin{tabular}{|rp{\pWidth}|}\hline\pStart&\\
\pRow{I will think step by step and answer your question.}
\pRow{}
\pRow{1. Question: Is growing seedless cucumber good for a gardener with entomophobia?}
\pRow{2. Explanation: Entomophobia is a fear of insects. Plants need insects to pollinate them. Seedless fruits such as seedless cucumbers do not require pollination so seedless fruits do not require insects. This is good for people with entomophobia.}
\pRow{3. Answer: Yes.}
\pRow{}
\pRow{1. Question: Who was British PM and Biceroy during Quit India Movement?}
\pRow{2. Explanation: The Quit India Movement was launched in 8th August 1942. The British PM at that time was Winston Churchill. The British Biceroy during the movement was Victor Hope, usually referred to as Lord Linlithgow.}
\pRow{3. Answer: Winston Churchil and Lord Linlithgow.}
\pRow{}
\pRow{1. Question: Which year does Game Over Man come out on Netflix?}
\pRow{2. Explanation: Game Over, Man! is an action-comedy movie. Its director is Kyle Newacheck. The movie stars Anders Holm, Adam DeVine, and Blake Anderson. The movie was released March 23, 2018 on Netflix.}
\pRow{3. Answer: March 23, 2018.}
\pRow{}
\pRow{1. Question: Would it be very difficult for Nuno Gomes to dive to the Red Sea's deepest point?}
\pRow{2. Explanation: Nuno Gomes' deepest dive in the Red Sea to date is 317 meters. The Red Sea has a maximum depth of over 3,000 meters. So it would be difficult for Nuno Gomes to dive to the deepest point of the Red Sea.}
\pRow{3. Answer: Yes.}
\pRow{}
\pRow{1. Question: Are chinchillas cold-blooded?}
\pRow{2. Explanation: Chinchillas are rodents. Rodents are mammals. All mammals are warm-blooded.}
\pRow{3. Answer: No.}
\pRow{}
\pRow{1. Question: What are the chemo drugs for gastric cancer?}
\pRow{2. Explanation: Chemotherapy can be given for gastric cancer before or after surgery. Usually you have a combination of 2 or 3 drugs. The most common types are: epirubicin, cisplatin and capecitabine (ECX) ; epirubicin, cisplatin and fluorouracil (ECF); or fluorouracil, folinic acid, oxaliplatin and docetaxel (FLOT).}
\pRow{3. Answer: Epirubicin, cisplatin and capecitabine (ECX) ; epirubicin, cisplatin and fluorouracil (ECF); or fluorouracil, folinic acid, oxaliplatin and docetaxel (FLOT).}
\pRow{}
\pRow{1. Question: How many sacks does Clay Matthews have in his career?}
\pRow{2. Explanation: Clay Matthews has been pro linebacker for 11 seasons. Ten of the seasons he played for Green Bay Packers, while his last season was with Los Angeles Rams, dedicated to his hometown. Clay Matthews III had 383 solo tackles, 136 assists, and 91.5 sacks. His father, Clay Matthews Jr., also played in NFL and had 69.5 sacks in his career.}
\pRow{3. Answer: 91.5 (Clay Matthews III) and 69.5 (Clay Matthews Jr.).}
\pRow{}
\pRow{1. Question: In the U.S., the most deaths are caused by?}
\pRow{2. Explanation: According to the Centers for Disease Control and Prevention, there were 3 million deaths in 2018. The leading cause of death in the U.S. is heart disease, causing nearly 700,000 deaths annually. The second most common cause of deaths is cancer.}
\pRow{3. Answer: Heart disease.}
\pRow{}
\pRow{1. Question: The rank of Indian economy in terms of nominal GDP is?}
\pRow{2. Explanation: The top GDP contries are United States, China, Japan, Germany, and India. India's GDP is economy is \$2.94 trillion. It is the fifth-largest economy in the world.}
\pRow{3. Answer: 5.}
\pRow{}
\pRow{1. Question: Who won season 16 on Dancing with the Stars?}
\pRow{2. Explanation: Dancing with the Stars is an American dance competition television series that premiered on June 1, 2005, on ABC. The 16th season premiered on March 18, 2013. It was won by Kellie Pickler.}
\pRow{3. Answer: Kellie Pickler.}
\pRow{}
\pRow{1. Question: \pVar{question}}
\pRow{2. Explanation: \pBlank}
&\\\hline\end{tabular}
\caption{The GPT-3 prompt for generating long-form answers to questions from NQ and SQA.}
\label{fig:prompt_passage_gpt3}
\end{center}
\end{figure*}

\begin{figure*}[p]
\begin{center}\scriptsize
\begin{tabular}{|rp{\pWidth}|}\hline\pStart&\\
\pRow{Hi, I will think step by step and answer your question.}
\pRow{Is growing seedless cucumber good for a gardener with entomophobia?}
\pRow{Yes. Entomophobia is a fear of insects. Plants need insects to pollinate them. Seedless cucumber fruit does not require pollination. So seedless cucumber does not require insects.}
\pRow{Here is another question. Which year does game over man come out on netflix?}
\pRow{Game Over, Man! is an American action-comedy movie. It was released on March 23, 2018, on Netflix. It is directed by Kyle Newacheck, written by Anders Holm, and starring himself, Adam DeVine, and Blake Anderson.}
\pRow{Here is another question. What are the chemo drugs for gastric cancer?}
\pRow{Chemotherapy can be given for gastric cancer before or after surgery. Usually you have a combination of 2 or 3 drugs. The most common types are: epirubicin, cisplatin and capecitabine (ECX) ; epirubicin, cisplatin and fluorouracil (ECF); or fluorouracil, folinic acid, oxaliplatin and docetaxel (FLOT).}
\pRow{Here is another question. Was N.W.A a album?}
\pRow{N.W.A was an American hip hop group from Compton, California.}
\pRow{Were they successful?}
\pRow{They has sold over 10 million units in the United States alone.}
\pRow{How many singles did it have?}
\pRow{The group NWA released 8 singles.}
\pRow{Here is another question. \pVar{$Q_1$}}
\pRow{\pVar{$A_1$}}
\pRow{...}
\pRow{\pVar{$Q_k$}}
\pRow{\pBlank}
&\\\hline\end{tabular}
\caption{The LaMDA prompt for generating answers to questions from QReCC. Each line is a conversation turn. The dialog context from QReCC contains rounds of questions and answers ($Q_1, A_1, Q_2, A_2, \dots, Q_k$).}
\label{fig:prompt_passage_lamda}
\end{center}
\end{figure*}

\begin{figure*}[p]
\begin{center}\scriptsize
\begin{tabular}{|rp{\pWidth}|}\hline\pStart&\\
\pRow{I will think step by step and answer your question.}
\pRow{}
\pRow{Is growing seedless cucumber good for a gardener with entomophobia?}
\pRow{Yes. Entomophobia is a fear of insects. Plants need insects to pollinate them. Seedless cucumber fruit does not require pollination. So seedless cucumber does not require insects.}
\pRow{}
\pRow{Which year does game over man come out on netflix?}
\pRow{Game Over, Man! is an American action-comedy movie. It was released on March 23, 2018, on Netflix. It is directed by Kyle Newacheck, written by Anders Holm, and starring himself, Adam DeVine, and Blake Anderson.}
\pRow{}
\pRow{What are the chemo drugs for gastric cancer?}
\pRow{Chemotherapy can be given for gastric cancer before or after surgery. Usually you have a combination of 2 or 3 drugs. The most common types are: epirubicin, cisplatin and capecitabine (ECX) ; epirubicin, cisplatin and fluorouracil (ECF); or fluorouracil, folinic acid, oxaliplatin and docetaxel (FLOT).}
\pRow{}
\pRow{Was N.W.A an album?}
\pRow{N.W.A was an American hip hop group from Compton, California.}
\pRow{Were they successful?}
\pRow{They has sold over 10 million units in the United States alone.}
\pRow{How many singles did they have?}
\pRow{N.W.A had eight singles, including "Straight Outta Compton", "Express Yourself", "Gangsta Gangsta", "Dopeman" and "Alwayz Into Somethin'".}
\pRow{}
\pRow{\pVar{$Q_1$}}
\pRow{\pVar{$A_1$}}
\pRow{...}
\pRow{\pVar{$Q_k$}}
\pRow{\pBlank}
&\\\hline\end{tabular}
\caption{The GPT-3 prompt for generating answers to questions from QReCC.}
\label{fig:prompt_passage_gpt3_conv}
\end{center}
\end{figure*}

} %